%% file: ms.tex
\documentclass[10pt,twocolumn,letterpaper]{article}

\usepackage{cvpr}              

\input{preamble}

\definecolor{cvprblue}{rgb}{0.21,0.49,0.74}
\usepackage[pagebackref,breaklinks,colorlinks,citecolor=cvprblue]{hyperref}

\def\paperID{8559} 
\def\confName{CVPR}
\def\confYear{2024}

\title{3DGazeNet: Generalizing 3D Gaze Estimation with Weak-Supervision \\ from Synthetic Views}

\author{
Evangelos Ververas\textsuperscript{1} \and
Polydefkis Gkagkos\textsuperscript{1} \and
Jiankang Deng\textsuperscript{1} \and
Michail Christos Doukas\textsuperscript{1,2} \and
Jia Guo\textsuperscript{3} \qquad
Stefanos Zafeiriou\textsuperscript{1,2}
\\
\textsuperscript{1}Huawei Technologies, London, UK
\textsuperscript{2}Imperial College London, UK
\textsuperscript{3}InsightFace
\\
{\tt\small \{evangelos.ververas, polydefkis.gkagkos, jiankang.deng, michail.christos.doukas,} 
\\
{\tt\small stefanos.zafeiriou\}@huawei.com} \qquad
{\tt\small guojia@gmail.com}
}

\begin{document}
\maketitle
\input{sec/0_abstract}    
\input{sec/1_intro}
\input{sec/2_relwork}
\input{sec/3_method}
\input{sec/4_experiments}
\input{sec/5_conclusion}

{
    \small
    \bibliographystyle{ieeenat_fullname}
    \bibliography{egbib}
}
\input{sec/X_suppl}


\end{document}

%% file: preamble.tex
\usepackage[dvipsnames]{xcolor}

\usepackage{multirow}

%% file: sec/0_abstract.tex
\begin{abstract}
Developing gaze estimation models that generalize well to unseen domains and in-the-wild conditions remains a challenge with no known best solution.
This is mostly due to the difficulty of acquiring ground truth data that cover the distribution of faces, head poses, and environments that exist in the real world.
Most recent methods attempt to close the gap between specific source and target domains using domain adaptation.
In this work, we propose to train general gaze estimation models which can be directly employed in novel environments without adaptation.
To do so, we leverage the observation that head, body, and hand pose estimation benefit from revising them as dense 3D coordinate prediction, and similarly express gaze estimation as regression of dense 3D eye meshes.
To close the gap between image domains, we create a large-scale dataset of diverse faces with gaze pseudo-annotations, which we extract based on the 3D geometry of the scene, and design a multi-view supervision framework to balance their effect during training.
We test our method in the task of gaze generalization, in which we demonstrate improvement of up to $30\%$ compared to state-of-the-art when no ground truth data are available, and up to $10\%$ when they are. 
The project material are available for research purposes at \href{https://github.com/Vagver/3DGazeNet}{https://github.com/Vagver/3DGazeNet}.
\end{abstract}

%% file: sec/1_intro.tex
\section{Introduction}
\label{sec:intro}

Eye gaze serves as a cue for understanding human behavior and intents, including attention, communication, and mental state.
As a result, gaze information has been exploited by a lot of applications of various fields of interest, ranging from medical and psychological analysis~\cite{Kleinke1986GazeAE,Castner2020,VIDAL20121306} to human-computer interaction~\cite{Sean2014}, efficient rendering in VR/AR headset systems~\cite{Meixu2020,Konrad2019,Burova2020}, virtual character animation~\cite{Richard_2021_WACV,Zhang2020dual,sun2022ide,sun2023next3d} and driver state monitoring~\cite{Mavely2017,kasahara2022look}. 

\begin{figure}
\setlength{\abovecaptionskip}{3pt}
\setlength{\belowcaptionskip}{3pt}
\centering
    \includegraphics[width=\linewidth]{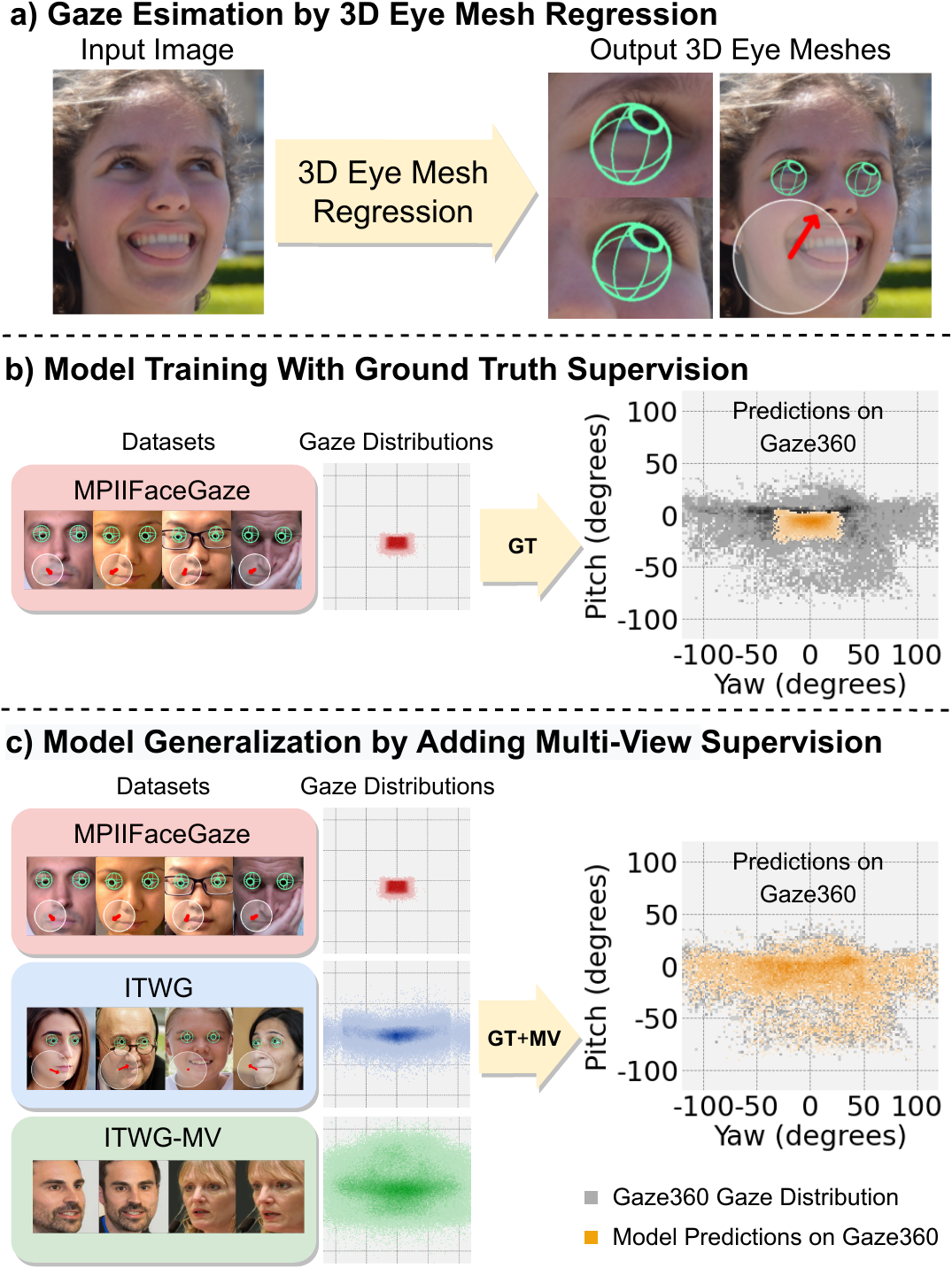}
    \caption{\small Overview of our method 3DGazeNet. a) We approach 3D gaze estimation as dense 3D eye mesh regression, which is robust against sparse prediction errors. b) Domain generalization is one of the hardest challenges in gaze estimation. Training with common gaze datasets often results in poor cross-dataset performance. c) Our multi-view supervision method employs pseudo-labels from in-the-wild face images to close the gap between controlled and in-the-wild datasets.}
\label{fig:summary}
\vspace{-10pt}
\end{figure}

Typically, 3D gaze estimation is expressed as a direct mapping between input images and a few pose parameters~\cite{zhang2017s_mpiifgaze,cvpr2016_gazecapture,Cheng_Huang_Wang_Qian_Lu_2020,Park2020ECCV_eve,wang2022contrastive}, or sparse representations of the eyes~\cite{wang2017real,Park2018ETRA,Park_2018_ECCV_dpge}.
Nevertheless, it has been shown that unconstrained face and body pose estimation from single images benefits from replacing predicting few pose or shape parameters by directly predicting dense 3D geometry~\cite{Deng_2020_CVPR,Kulon_2020_CVPR,Guler2018DensePose,Guler_2019_CVPR,alp2017densereg}.
In this work, we leverage this observation and revise the formulation of gaze estimation as end-to-end dense 3D eye mesh regression, which combined with standard vector regression induces multiple benefits.
Existing datasets with ground truth 3D eyes include only images in the IR domain~\cite{Fuhl2021TEyeDO2}, however, IR images cannot be directly employed for RGB-based methods.
As 3D eye meshes are not available for most gaze datasets, we define a unified eye representation, i.e. a rigid 3D eyeball template (\cref{fig:eyeball}(a)), which we fit on images based on sparse landmarks and the available gaze labels.

Several gaze datasets have become available in the last decade~\cite{CAVE_Columbia2013,Park2020ECCV_eve,Zhang2020ETHXGaze,FunesMora_ETRA_2014_eyediap,zhang15_cvpr_mpiigaze,cvpr2016_gazecapture,gaze360_2019,sugano2014utmv,Fischer_2018_ECCV}, which have contributed to the recent progress in automatic 3D gaze estimation from monocular RGB images.
However, collecting gaze datasets is a costly and challenging process which often restricts them being captured in controlled environments and consisting of limited unique identities, thus lacking variation compared to data from the real world.
This causes the most common challenge in gaze estimation, which is cross-domain and in-the-wild generalization. 
In this work, we propose a method to exploit arbitrary, unlabeled face images to largely increase the diversity of our training data as well as our model's generalization capabilities.
To that end, we design a simple pipeline to extract robust 3D gaze pseudo-labels based on the 3D shape of the face and eyes, without having any prior gaze information.
Based on recent advancements on weakly-supervised head, body and hand pose estimation~\cite{Cai_2018_ECCV,deng2019accurate,Iqbal_2020_CVPR,Li_Li_Jiang_Zhang_Huang_Xu_2020,Wandt2021Canonpose}, we regularize inconsistencies of pseudo-labels, by a geometric constraint which encourages our model to maintain prediction consistency between multiple synthetic views of the same subject.

Most recent methods attempt to close the gap between diverse image domains using domain adaptation. 
Commonly, they employ a few samples of the target domain, with \cite{Park2019ICCV_fewshot,Yu_2019_CVPR,He_2019_ICCVW} or without \cite{Wang_2019_CVPR_bayesian,Guo_2020_ACCV,Liu_2021_ICCV,wang2022contrastive,bao2022generalizing,cheng2022puregaze,ghosh2022mtgls,cai2023source} their gaze labels, to fine-tune an initial model. 
Although successful, approaches following this scheme require knowledge of the target domain and model re-training, which prohibit their use as plug-n-play methods in real user applications.
In contrast, we propose a method to train gaze estimation models that generalize well to unseen and in-the-wild environments without the constraints of domain adaption. Our method can effortlessly be employed by user applications in a plug-n-play fashion.

An overview of our approach, which we name 3DGazeNet, is presented in \cref{fig:summary}.
We evaluate our method in cross-dataset gaze generalization, showcasing improvements over the state-of-the-art, even by a large margin, and perform ablations over the model components.
To summarize, the key contributions of our work are:
\begin{itemize}
    \item A simple automatic method to extract robust 3D eye meshes from arbitrary face images and a multi-view consistency regularization which allows to exploit them for improved gaze generalization. 
    \item A revised formulation for gaze estimation, based on dense 3D eye mesh regression from images. To the best of our knowledge, we are the first to utilize an end-to-end 3D eye mesh regression approach for gaze estimation.
    \item Improved performance over the state-of-the-art in gaze generalization with ($10\%$) and without ($30\%$) using source domain ground truth, with a simple model architecture. Based on that, we believe that 3DGazeNet is an important step towards reliable plug-n-play gaze tracking.
\end{itemize}

%% file: sec/2_relwork.tex
\section{Related Work}
\label{sec:related_work}

Numerous model designs for supervised 3D gaze estimation have been tested recently, investigating which face region to use as input~\cite{zhang2017s_mpiifgaze,cvpr2016_gazecapture,Cheng_Huang_Wang_Qian_Lu_2020}, the model architecture~\cite{liu2018differential,Wang_2018_CVPR,Cheng_2018_ECCV} and what external stimuli to utilize to improve performance~\cite{Park2020ECCV_eve}. 
Motivated by the difficulties in collecting diverse and large scale data for gaze estimation, recent works have shown that valuable gaze representations can be extracted in fully unsupervised settings, by applying gaze redirection~\cite{Yu_2020_CVPR} or disentanglement constraints~\cite{Sun_2021_ICCV}.

\smallskip
\noindent\textbf{Gaze Adaptation and Generalization}
Much effort has also been made to design methods that adapt well to known target subjects and environments, by employing either few labeled samples~\cite{Park2019ICCV_fewshot,Yu_2019_CVPR,He_2019_ICCVW} or completely unlabeled data of the target domain~\cite{Wang_2019_CVPR_bayesian,Guo_2020_ACCV,Liu_2021_ICCV,wang2022contrastive,bao2022generalizing,cheng2022puregaze,ghosh2022mtgls,cai2023source}.
Differently from the above, gaze generalization models aim to improve cross-domain performance without any knowledge of the target domains. 
The models in~\cite{wang2022contrastive,bao2022generalizing,cheng2022puregaze}, even though targeted for gaze adaptation, are based on learning general features for gaze estimation and thus, they perform well in target domain-agnostic settings.
Moreover, \cite{Kothari_2021_CVPR} has shown that it is possible to train general gaze estimation models by employing geometric constraints in scenes depicting social interaction between people.
We believe that~\cite{Kothari_2021_CVPR} is the closest work to ours, as it is the only method which uses 3D geometric cues of the scene to learn gaze from arbitrary face data.
Lastly,~\cite{zhang2022gazeonce} proposes to improve generalization by employing synthetic images which are, however, limited by the gaze distribution of existing gaze datasets.
Both the implementation and custom dataset are not public, which hinders reproducibility and reliable comparisons.

\begin{figure}
    \setlength{\tabcolsep}{3pt}
    \renewcommand{\arraystretch}{0.}
    \centering
    \begin{tabular}{c}
        \includegraphics[width=0.47\textwidth]{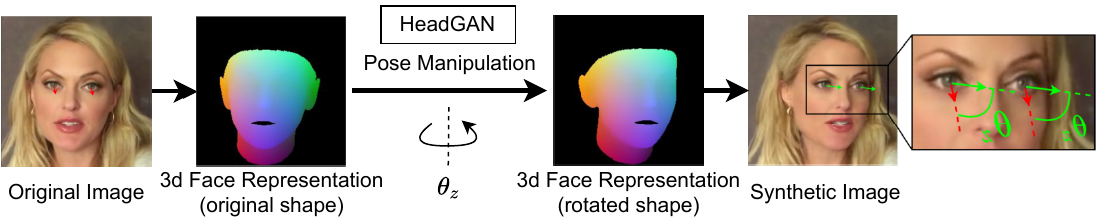}
        \\ \small (a) \\
        \includegraphics[width=0.45\textwidth]{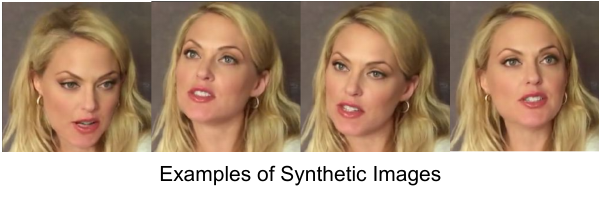}
        \\ \small (b)
    \end{tabular}
    \vspace{-10pt}
    \caption{\small (a) We use HeadGAN~\cite{doukas2020headgan} to generate novel views by manipulating the 3D pose of the face. During synthesis, angle $\theta_z$ is transferred to all facial parts including the eyes, thus the relative angle between the head and eyes (i.e. the gaze direction in the head coordinate system) is maintained. (b) Images generated with HeadGAN by rotating the face depicted in the original image.}
    \label{fig:headgan}
\vspace{-10pt}
\end{figure}

\smallskip
\noindent\textbf{Model-Based Gaze Estimation} Differently from the above, sparse or semantic representations of the eye geometry have also been employed by some methods to infer gaze from images~\cite{3dmm_wood2016,wang2017real,Park2018ETRA,Yu_2018_ECCVW,Park_2018_ECCV_dpge,Wang_2018_CVPR}.
However, such representations do not convey information about the 3D substance of eyes and are prone to noisy predictions.
In contrast, by predicting 3D eye meshes we learn a much more robust representation, from which we can retrieve any other sparse or semantic one just by indexing.
Recovering dense 3D geometry of the eye region from images by fitting parametric models of the shape and texture has been previously proposed~\cite{3dmm_wood2016}.
However, restrictions posed by building large-scale parametric models and fitting in-the-wild images have resulted in low gaze accuracy compared to learning-based methods. 

\smallskip
\noindent\textbf{Face Reenactment and Learning from Synthetic Data} Synthetic image data have been previously used in training deep networks, mainly for data augmentation and pseudo-ground truth generation. For instance, \cite{10.1007/978-3-319-93040-4_28} used CycleGAN~\cite{CycleGAN2017} to create a new training corpus in order to balance emotion classes in the task of emotion classification.
More recently, GANcraft~\cite{hao2021GANcraft} employed SPADE~\cite{park2019SPADE} to generate pseudo-ground truth images that were used to supervise their neural rendering framework.
In this work, we obtain access to image pairs of the same subject in different views, by taking advantage of HeadGAN~\cite{doukas2020headgan}, a face reenactment system.
In contrast to person-specific reenactment methods~\cite{kim2018deep,head2head2020,head2headpp} or person-generic landmark-driven approaches~\cite{Zakharov2019FewShotAL,bilayer,fsvid2vid}, HeadGAN is able to perform free-view synthesis using a single source image.


%% file: sec/3_method.tex
\section{Method}
\label{sec:method}

\subsection{Problem Definition and Motivation}
\label{sec:motivation_headgan}


The aim of this work is to design a method that given a face image $\mathbf{I}$, it estimates $2\times N_v$ 3D coordinates $\mathbf{V} = [\mathbf{V}_{l}^T, \mathbf{V}_{r}^T]^T$, where $\mathbf{V}_{l} \in \mathbb{R}^{N_v\times3}$ are coordinates corresponding to the left eyeball while $\mathbf{V}_{r} \in \mathbb{R}^{N_v\times3}$ to the right, as well as a 3D gaze vector $g=(g_x,g_y,g_z)$.
Then, the final gaze result is calculated by the mean direction of the two output components.
Inspired by recent work in self-supervised 3D body pose estimation~\cite{Li_Li_Jiang_Zhang_Huang_Xu_2020,Iqbal_2020_CVPR,Wandt2021Canonpose}, we adopt multi-view constraints to train our model based on in-the-wild faces and automatically generated gaze pseudo-labels. 

To employ multi-view losses, we assume that images of the same subject with different head poses and the same gaze direction relatively to the head are available.
For example, this condition is satisfied when a face picture is taken from different angles at the same time.
As such images are not commonly available for in-the-wild datasets, we employ HeadGAN~\cite{doukas2020headgan}, a recent face reenactment method, to generate novel face poses from existing images.
HeadGAN is able to synthesize face animations, using dense face geometry, which covers the eyes, as a driving signal and single source images.
Using dense geometry guarantees that the relative angle between the head and eyes is maintained when synthesizing novel poses, as it is shown in \cref{fig:headgan}.

\subsection{Unified 3D Eye Representation}
\label{sec:unified_representation}

\begin{figure}
    \setlength{\tabcolsep}{1pt}
    \renewcommand{\arraystretch}{1.}
    \centering
    \begin{tabular}{c}
        \includegraphics[width=0.7\linewidth]{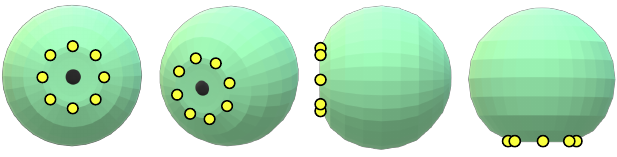}
        \\
        \small $\mathbf{M}$: $N_v = 481$ vertices, $N_t = 928$ triangles
        \\
        \small (a) Eyeball template
        \\
        \includegraphics[width=0.95\linewidth]{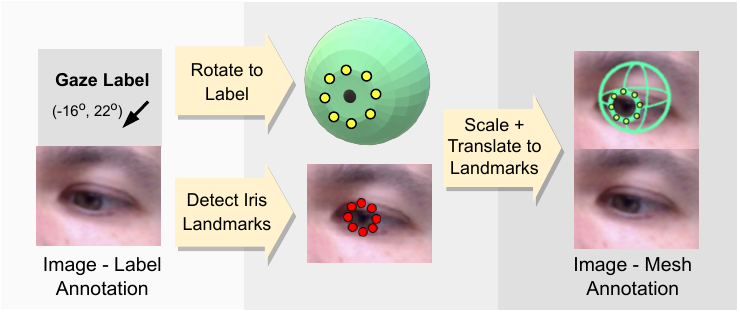}
        \\
        \small (b) Ground truth generation
        \\
        \includegraphics[width=0.95\linewidth]{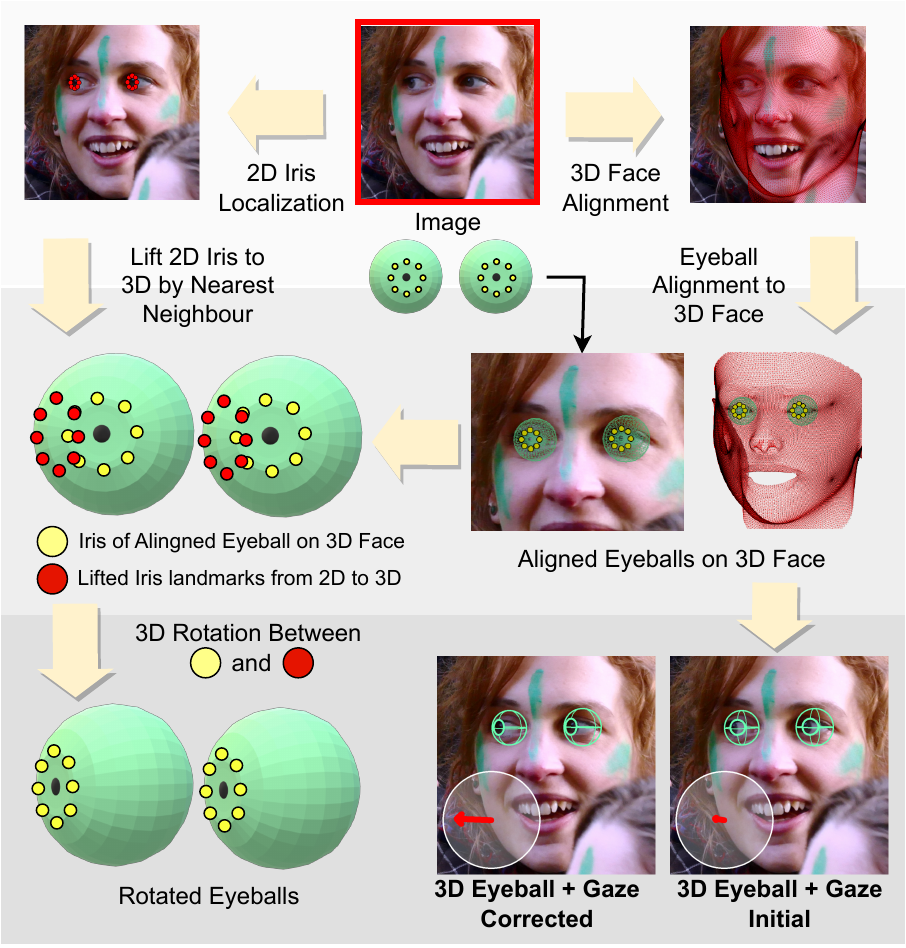}
        \\
        \small (c) Pseudo-ground truth generation
    \end{tabular}
    \caption{\small (a) The employed rigid 3D eyeball mesh template. (b) Ground truth data generation, applied on gaze estimation datasets with available ground truth. (c) Pseudo-ground truth data generation, applied on arbitrary face images without any gaze label.}
    \label{fig:eyeball}
\vspace{-10pt}
\end{figure}

Learning consistent eye meshes across different images and datasets, requires establishing a unified 3D eye representation.
To that end, we define a 3D eyeball template as a rigid 3D triangular mesh with spherical shape, consisting of $N_v = 481$ vertices and $N_t = 928$ triangles.
We create two mirrored versions, $\mathbf{M}_l$ and $\mathbf{M}_r$, of the above mesh to represent a left and a right reference eyeball respectively.
This representation allows us to allocate semantic labels to specific vertices of the eyeball, such as the iris border (\cref{fig:eyeball} (a)), and calculate 3D gaze direction as the orientation of the central axis of our 3D eyeball template.
In practice, an offset angle (the kappa coefficient) exists between the central (optical) and visual axes of eyes, which is subject-dependent and varies between $-2^o$ to $2^o$ across the population~\cite{Yu_2019_CVPR}.
Accounting for this offset is essential for person-specific gaze estimation~\cite{He_2019_ICCVW,Liu2018ADA,Park2019ICCV_fewshot,Yu_2019_CVPR}.
However, in our case of cross-dataset and in-the-wild gaze generalization, in which errors are much larger than the possible offset, data diversity is more important than anatomical precision and thus, our spherical eyeball is a reasonable approximation.

\smallskip
\noindent{\textbf{3D Eyes Ground-Truth from Gaze Datasets}}
For gaze estimation datasets, exact supervision can be acquired by automatically fitting the eyeball template on face images, based on sparse iris landmarks and the available gaze labels, as shown in \cref{fig:eyeball}(b).
Specifically, we first rotate the eyeball template around its center according to the gaze label. 
Then, we align (scale and translation) $x$, $y$ coordinates of the rotated eye mesh to the iris landmarks of the image and multiply $z$ coordinates with the same scale.
To extract sparse iris landmarks we employed the method of~\cite{Park2018ETRA}, but any similar method could have been used.

\smallskip
\noindent{\textbf{3D Eyes Pseudo-Ground Truth from In-The-Wild Images}}
To extract 3D eyes from images without gaze labels, we have developed an automatic pipeline based on 3D face alignment and 2D iris localization.
Having recovered the 3D face with $x$, $y$ in image space, we first align our eyeball templates in the eye sockets based on the face's eyelid landmarks.
Then, we lift 2D iris predictions to 3D by finding the nearest vertexes from the aligned 3D eye templates.
Finally, we compute the rotation between the initially aligned eyes and the 3D-lifted iris center and rotate the eyeballs accordingly.
For 3D face alignment, we employ RetinaFace~\cite{Deng_2020_CVPR} and for 2D iris localization~\cite{Park2018ETRA} as above. The process is presented in \cref{fig:eyeball}(c).

\subsection{Joint 3D Eye Mesh and Vector Regression}
\label{sec:3deye_reg}

Given an input face image $\mathbf{I}$, we utilize 5 face detection landmarks to crop patches around each one of the two eyes.
We resize the patches to shape $128 \times 128 \times 3$ and stack them channel-wise along with a cropped image of the face.
We employ a simple model architecture consisting of a ResNet-18~\cite{he2016deep} to extract features, followed by two fully connected layers to map features to two separate eye modalities, which are a) dense 3D eye coordinates and b) a 3D gaze vector.
As the final gaze output, we consider the mean direction calculated from the two modalities.


To train the above network for mesh regression, similarly to \cite{Deng_2020_CVPR}, we enforce a vertex loss and an edge length loss between the model outputs and the respective ground truth or pseudo-ground truth, which can be expressed as:
\begin{equation}
    \setlength{\abovedisplayskip}{3pt}
    \setlength{\belowdisplayskip}{5pt}
    \begin{split}
        \mathcal{L}_{vert} &= 
        \frac{1}{N_v} \sum_{j=\{l,r\}} \sum_{i=1}^{N_v} \| \mathbf{V}_{j,i} - \mathbf{V}^*_{j,i} \|_1,
        \label{eq:vertex_loss}
    \end{split}
\end{equation}
where $\mathbf{V}_{j}\in\mathbb{R}^{N_v\times 3}$ and $\mathbf{V}^*_{j}\in\mathbb{R}^{N_v\times 3}$  for $j=\{l,r\}$ are the output and the (pseudo-)ground truth coordinates, while the edge length loss (based on the fixed mesh triangulation of our template meshes) can be written as:
\begin{equation}
    \setlength{\abovedisplayskip}{3pt}
    \setlength{\belowdisplayskip}{5pt}
    \begin{split}
        \mathcal{L}_{edge} &= 
        \frac{1}{3{N_t}} \sum_{j=\{l,r\}} \sum_{i=1}^{3N_t} \| \mathbf{E}_{j,i} - \mathbf{E}^*_{j,i} \|_2
        \label{eq:edge_loss}
    \end{split}
\end{equation}
where $\mathbf{E}_{j}\in\mathbb{R}^{3N_t}$ and $\mathbf{E}^*_{j}\in\mathbb{R}^{3N_t}$ for $j=\{l,r\}$ are the edge lengths of the predicted and the (pseudo-)ground truth eyes. As edge length we define the Euclidean distance between two vertices of the same triangle. In addition to the mesh regression losses, we enforce a gaze loss to the gaze output of our model, expressed as:
\begin{equation}
    \setlength{\abovedisplayskip}{3pt}
    \setlength{\belowdisplayskip}{5pt}
    \begin{split}
        \mathcal{L}_{gaze} = (180 / \pi) \arccos(\mathbf{g}^T\mathbf{g}^*)
        \label{eq:gaze_loss}
    \end{split}
\end{equation}
where $\mathbf{g}$ and $\mathbf{g}^*$ are the normalized model output and the gaze (pseudo-)ground truth respectively. We combine losses of \cref{eq:vertex_loss,eq:edge_loss,eq:gaze_loss} in a single loss function to train our models with supervision from (pseudo-)ground truth 3D eye meshes and gaze vectors. The combined loss is written as:
\begin{equation}
    \setlength{\abovedisplayskip}{3pt}
    \setlength{\belowdisplayskip}{3pt}
    \begin{split}
        \mathcal{L}_{(P)GT} = \lambda_v\mathcal{L}_{vert} + \lambda_e\mathcal{L}_{edge} + \lambda_g\mathcal{L}_{gaze},
        \label{eq:gt_loss}
    \end{split}
\end{equation}
where $\lambda_v$, $\lambda_e$, and $\lambda_g$ are parameters which regularize the contribution of the loss terms in the overall loss. From our experiments we have selected their values to be $\lambda_v=0.1$, $\lambda_e=0.01$ and $\lambda_g=1$.

\begin{figure*}
    \centering
    \includegraphics[width=0.9\textwidth]{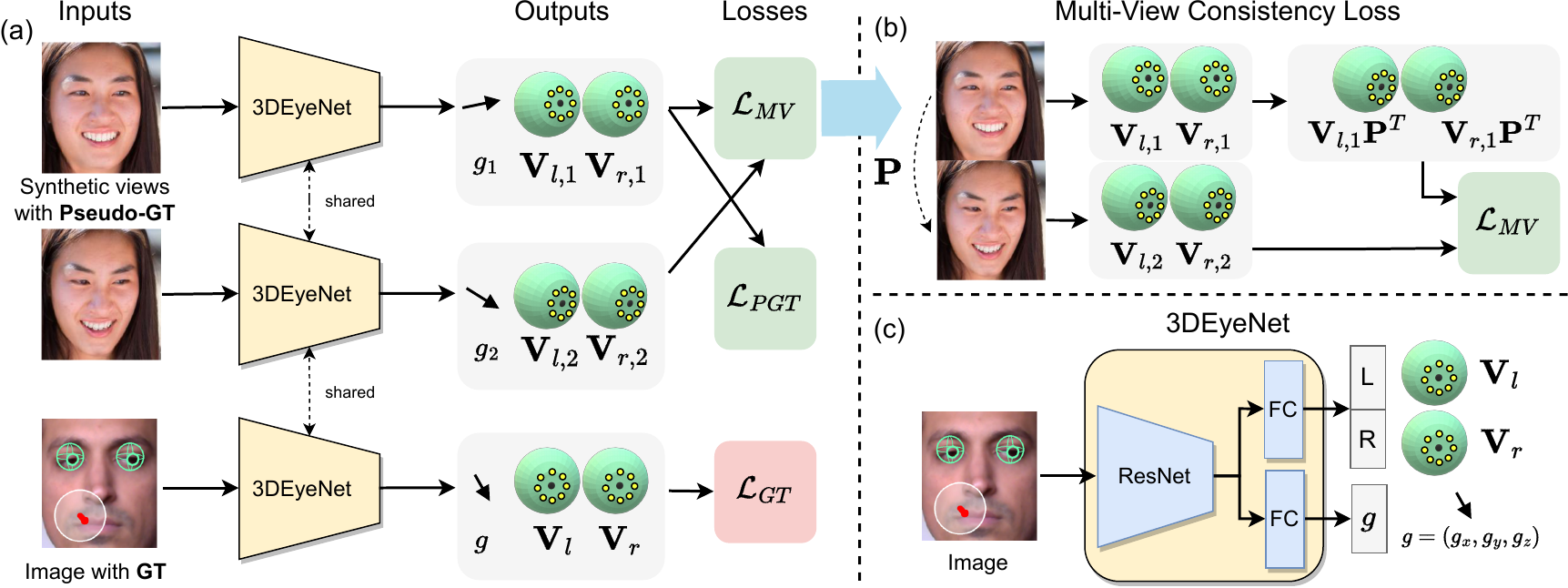}
    \caption{\small Overview of the proposed method 3DGazeNet. a) During training we employ single images with ground-truth supervision or pairs of synthetic views of the same subject with pseudo-annotations and different head poses. Different sets of losses are employed depending on the type of supervision. b) Detailed demonstration of $\mathcal{L}_{MV}$. 3D transformation $\mathbf{P}$ which maps view 1 to view 2, is employed to transform points $\mathbf{V}_{l,1}$ and $\mathbf{V}_{r,1}$, before calculating an L1 distance loss against $\mathbf{V}_{l,2}$ and $\mathbf{V}_{r,2}$. c) The base network (3DEyeNet) of our model consists of a ResNet-18 backbone and two fully connected layers leading to the 3D eye mesh and gaze vector outputs.}
    \label{fig:main}
\vspace{-10pt}
\end{figure*}

\subsection{Multi-View Consistency Supervision}
\label{sec:multi_view}

Extending our training dataset with in-the-wild images and training using pseudo-ground truth, usually improves the ability of models to generalize to unseen domains, as can be seen by our experiments in \cref{sec:ablation}.
However, automatically generated 3D eyes and gaze include inconsistencies which are hard to identify and filter out.
To balance the feedback of direct supervision from pseudo-ground truth, we design a multi-view supervision framework, based on pairs of real and synthetic images with different head poses, generated by HeadGAN as described in \cref{sec:motivation_headgan}.

Recovering dense 3D face coordinates and pose from images has recently been quite reliable \cite{Deng_2020_CVPR,Deng_2020_CVPR,Gecer_2019_CVPR,albiero2021img2pose}.
Having a pair of images $\mathbf{I}_1$ and $\mathbf{I}_2$ of the same subject and their reconstructed 3D faces, we can compute a transformation matrix $\mathbf{P} \in \mathbb{R}^{3\times4}$ which aligns the two faces in image space.
Assuming that gaze direction in both images remains still relative to the face, as is the case with images created by HeadGAN, we are able to supervise 3D regression of eyes by restricting our model's predictions to be consistent over an image pair, as output vertices should coincide when transformation $\mathbf{P}$ is applied to one of the pair's outputs. A similar approach has been employed successfully for weakly-supervised body pose estimation~\cite{Li_Li_Jiang_Zhang_Huang_Xu_2020,Iqbal_2020_CVPR,Wandt2021Canonpose}.
Particularly, we form a pair vertex loss as:
\begin{equation}
    \setlength{\abovedisplayskip}{3pt}
    \setlength{\belowdisplayskip}{5pt}
    \mathcal{L}_{MV,vertex} = \frac{1}{N_v} \sum_{j=\{l,r\}} \sum_{i=1}^{N_v} \| \mathbf{V}_{1,j,i}\mathbf{P}^T - \mathbf{V}_{2,j,i} \|_1,
    \label{eq:pair_vert_loss}
\end{equation}
where $\mathbf{V}_{1,j}, \mathbf{V}_{2,j} \in\mathbb{R}^{N_v \times 4}$ for $j=\{l,r\}$ are the output matrices for left and right eyes, which correspond to input images $\mathbf{I}_1$ and $\mathbf{I}_2$. $\mathbf{V}_{1,j,i}, \mathbf{V}_{2,j,i} \in \mathbb{R}^4$ are the specific homogeneous 3D coordinates indexed by $i$ in the above matrices. 
To enforce consistency constraints to the gaze head of our model, we analyse matrix $\mathbf{P}$ to scale $s$, rotation $\mathbf{R}$ and translation $\mathbf{t}$ components and employ $\mathbf{R}$ in a gaze loss:
\begin{equation}
    \setlength{\abovedisplayskip}{3pt}
    \setlength{\belowdisplayskip}{5pt}
    \mathcal{L}_{MV,gaze} = (180 / \pi) \arccos((\mathbf{g}^T_1\mathbf{R}^T)\mathbf{g}_2)
    \label{eq:pair_gaze_loss}
\end{equation}
where $\mathbf{g}_1$ and $\mathbf{g}_2$ are the normalized model outputs for input images $\mathbf{I}_1$ and $\mathbf{I}_2$ respectively. We combine losses of \cref{eq:pair_vert_loss,eq:pair_gaze_loss} in a single loss function to enforce multi-view consistency in mesh and gaze vector regression, between model outputs coming from pairs of input images. The combined loss is written as:
\begin{equation}
    \setlength{\abovedisplayskip}{3pt}
    \setlength{\belowdisplayskip}{3pt}
    \begin{split}
        \mathcal{L}_{MV} = \lambda_{MV,v}\mathcal{L}_{MV,vertex} + \lambda_{MV,g}\mathcal{L}_{MV,gaze},
        \label{eq:mv_loss}
    \end{split}
\end{equation}
where $\lambda_{MV,v}$ and $\lambda_{MV,g}$ are parameters which regularize the contribution of the loss terms in the overall loss. In our experiments, we have selected their values to be $\lambda_{MV,v} = 0.1$ and $\lambda_{MV,g} = 1$. 
To train models with all supervision signals, i.e. ground truth ($\mathcal{L}_{GT}$), pseudo-ground truth ($\mathcal{L}_{PGT}$) and multi-view supervision ($\mathcal{L}_{MV}$), we utilize the following overall loss function:
\begin{equation}
    \setlength{\abovedisplayskip}{3pt}
    \setlength{\belowdisplayskip}{3pt}
    \begin{split}
        \mathcal{L} = \lambda_{GT}\mathcal{L}_{GT} + \lambda_{PGT}\mathcal{L}_{PGT} + \lambda_{MV}\mathcal{L}_{MV},
        \label{eq:full_loss}
    \end{split}
\end{equation}
with parameters $\lambda_{GT}$, $\lambda_{PGT}$ and $\lambda_{MV}$ set to one. 
Implementation details are included in the supplementary material. 
An overview of 3DGazeNet is presented in \cref{fig:main}.

%% file: sec/4_experiments.tex
\section{Experiments}
\label{sec:experiments}


\subsection{Datasets}
\label{sec:datasets}

\begin{figure}
    \setlength{\tabcolsep}{-1.5pt}
    \renewcommand{\arraystretch}{0.5}
    \centering
    \includegraphics[width=\linewidth]{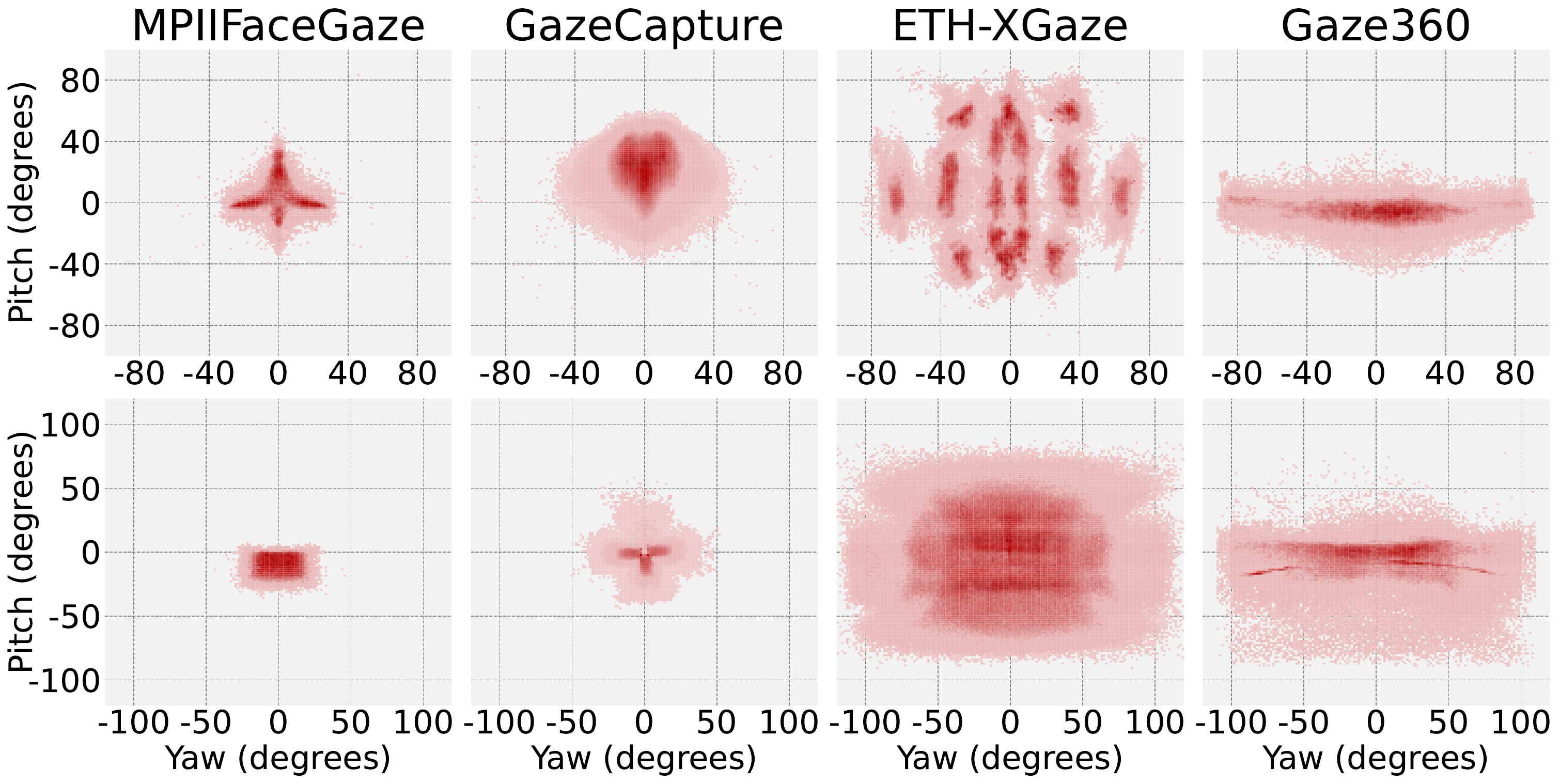}
    \includegraphics[width=\linewidth]{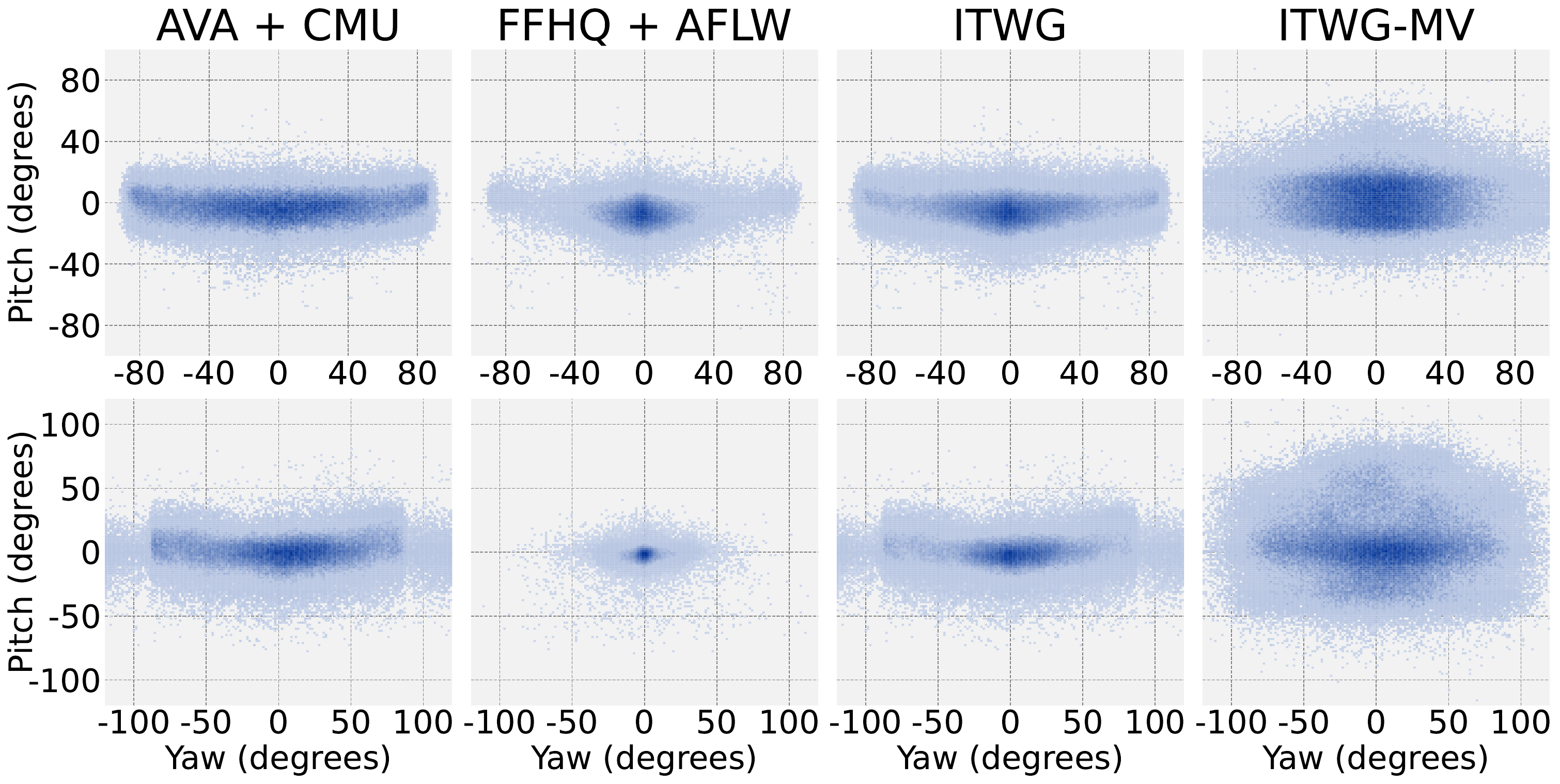}
    \caption{\small Distributions of the head pose (lines 1 and 3) and gaze (lines 2 and 4) of the employed datasets. Wide distribution datasets CMU, AVA, FFHQ, and AFLW are exploited to close the gap between diverse image domains.}
    \label{fig:data_distributions}
\end{figure}

\begin{table*}
\setlength{\tabcolsep}{2pt}
\small
\caption{\small Weakly-supervised method evaluation in cross- and within-dataset experiments. In all cases, we calculate gaze error in degrees (lower is better), on the test set of Gaze360. CMU and AVA correspond to subsets of ITWG-MV (i.e. augmented for multi-view supervision), providing a clearer comparison with~\cite{Kothari_2021_CVPR}. Our method trained with ITWG-MV outperforms the baselines in all cases.}
\label{tab:weak}
\centering
\resizebox{0.99\linewidth}{!}{
\begin{tabular}{c|c|c|c}
    \multicolumn{1}{c}{\footnotesize\textbf{(a) Cross-dataset}} &
    \multicolumn{2}{c}{\footnotesize\textbf{(b) Cross-dataset}} &
    \multicolumn{1}{c}{\footnotesize\textbf{(c) Within-dataset}}
    \\
    \multicolumn{1}{c}{\footnotesize\textbf{Synthetic Views}}        & 
    \multicolumn{2}{c}{\footnotesize\textbf{Ground Truth + Synthetic Views}} &
    \multicolumn{1}{c}{\footnotesize\textbf{Ground Truth + Synthetic Views}}
    \\
    \toprule
    \begin{tabular}[t]{lcc}
        \multicolumn{1}{c}{Dataset} & \cite{Kothari_2021_CVPR} & 3DGazeNet
        \\
        \midrule
        AVA     & 29.0 & \textbf{22.4} \\
        CMU     & 26.0 & \textbf{20.3} \\
        CMU+AVA & 22.5 & \textbf{19.7} \\
        ITWG-MV & -    & \textbf{18.1} 
    \end{tabular}
    &
    \begin{tabular}[t]{lccc}
        \multicolumn{1}{c}{Dataset} & \cite{Zhang2020ETHXGaze} & \cite{Kothari_2021_CVPR} & 3DGazeNet
        \\
        \midrule
        GC         & 30.2 & 29.2 & \textbf{27.5} \\
        GC+AVA     & -    & 19.5 & \textbf{18.9} \\
        GC+AVA+CMU & -    & -    & \textbf{18.4} \\
        GC+ITWG-MV & -    & -    & \textbf{17.6} 
    \end{tabular}
    &
    \begin{tabular}[t]{lccc}
        \multicolumn{1}{c}{Dataset} & \cite{Zhang2020ETHXGaze} & \cite{Kothari_2021_CVPR} & 3DGazeNet
        \\
        \midrule
        EXG         & 27.3 & 20.5 & 22.1 \\
        EXG+AVA     & -    & 16.9 & 17.1 \\
        EXG+AVA+CMU & -    & -    & \textbf{16.7} \\
        EXG+ITWG-MV & -    & -    & \textbf{15.4} 
    \end{tabular}
    &
    \begin{tabular}[t]{lccc}
        \multicolumn{1}{c}{Dataset} & \cite{gaze360_2019} & \cite{Kothari_2021_CVPR} & 3DGazeNet
        \\
        \midrule
        G360         & 11.1 & 10.1 & \textbf{9.6} \\
        G360+AVA     & -    & 10.2 & \textbf{9.7} \\
        G360+AVA+CMU & -    & -    & \textbf{9.5} \\
        G360+ITWG-MV & -    & -    & \textbf{9.3}
    \end{tabular}
    \\
    \bottomrule
\end{tabular}
}
\end{table*}

\noindent\textbf{Gaze Datasets}
Captured in a lab environment, ETH-XGaze (EXG)~\cite{Zhang2020ETHXGaze} consists of 756K frames of 80 subjects and includes large head pose and gaze variation.
Collected in uncontrolled indoor environments with mobile devices, MPIIFaceGaze (MPII)~\cite{zhang15_cvpr_mpiigaze} includes smaller head pose and gaze variation and consists of 45K images of 15 subjects, while GazeCapture (GC)~\cite{cvpr2016_gazecapture} contains almost 2M frontal face images of 1474 subjects.
In contrast to the above datasets, Gaze360 (G360)~\cite{gaze360_2019} is the only gaze dataset captured both indoors and outdoors and consists of 127K training sequences from 365 subjects.
The large variation in head pose, gaze, and environmental conditions makes Gaze360 the most challenging yet appropriate benchmark for in-the-wild gaze estimation.
For our experiments, we normalized the above datasets based on~\cite{zhang2018revisiting}, except for Gaze360 which we process to get normalized face crops. Additionally, we employ the predefined training-test splits, while for Gaze360 we only use the frontal facing images with head pose yaw angle up to $90^o$.
The head pose and gaze distributions of the above datasets are presented in \cref{fig:data_distributions}.

\smallskip
\noindent\textbf{In-The-Wild Face Datasets}
In-the-wild face datasets consist of significantly more unique subjects and capturing environments.
For our experiments, we employed four publicly-available datasets FFHQ~\cite{Karras_2019_CVPR} (70K images), AFLW~\cite{koestinger2011annotated} (25K images), AVA~\cite{Gu_2018_CVPR,Marin-Jimenez_2019_CVPR,marin21pami} and CMU-Panoptic~\cite{cmu2019}. 
FFHQ and AFLW are in-the-wild face datasets commonly used for face analysis, AVA is a large-scale in-the-wild human activity dataset annotated under the Looking-At-Each-Other condition and CMU-Panoptic is collected in lab conditions and captures interactions of multiple people in the same scene. 

FFHQ and AFLW include one face per image and thus are only processed to get normalized face crops.
AVA and CMU-Panoptic include frames with multiple faces, from which we randomly select 80K faces from each dataset with a maximum head pose of $90^o$.
Similarly to~\cite{Kothari_2021_CVPR}, for CMU we employed only frames captured with cameras in eye height.
We name this collection of 255K images as the ``In-The-Wild Gaze'' dataset (ITWG).
Lastly, to enforce multi-view supervision as described in \cref{sec:multi_view}, we synthesized novel views from images of ITWG using HeadGAN, sampling the pitch and yaw angles from Gaussians $\mathcal{N}(0, 20)$, relatively to the original head pose.
We name this collection of images as ``Multi-View In-The-Wild Gaze'' dataset (ITWG-MV) and employ it in our experiments to improve the generalization of gaze estimation.
The head pose and gaze distributions of the above datasets are presented in \cref{fig:data_distributions}.

\subsection{Gaze Generalization}
\label{sec:mv_eval}

In this section, we evaluate 3DGazeNet in within-dataset and cross-dataset experiments. We believe that~\cite{Kothari_2021_CVPR} is the most closely related method to ours, as it is the only method using 3D geometric cues of the scene to generalize gaze from arbitrary face data.

\smallskip
\noindent{\bf Cross-dataset Evaluation}
We design two cross-dataset experiments to test the generalization of our method on G360 and report the results on \cref{tab:weak}(a) and (b).
Particularly, the experiments are: 
a) we train our method on the CMU, AVA, and ITWG-MV datasets utilizing only our pseudo-labels and multi-view supervision and 
b) we additionally employ ground truth supervision from GC and EXG. 
From the results of the above experiments, it becomes obvious that our geometry-aware pseudo-labels employed within our multi-view supervision training effectively generalize gaze estimation to unseen domains, even without any available ground truth.
In particular, in experiment a) our method outperforms~\cite{Kothari_2021_CVPR} by 30\% with AVA, 12\% with CMU, 9.5\% with AVA+CMU and 14\% with our large-scale ITWG-MV.
Similarly, in experiment b) when additionally employing ground truth from GC and EXG our method outperforms~\cite{Kothari_2021_CVPR} by 7\% and 6\% respectively.

\noindent{\bf Within-dataset Evaluation}
Here we compare our method against state-of-the-art within-dataset gaze estimation on G360. Similarly to~\cite{Kothari_2021_CVPR}, we employ AVA for additional supervision, while we also examine the effect of the larger-scale ITWG-MV. The results, presented in \cref{tab:weak} (c), show that multi-view supervision from AVA does not improve performance (which is in line with the compared method), but the large-scale ITWG-MV does. 

\smallskip
\noindent{\bf Comparison with state-of-the-art} We further compare 3DGazeNet against recent methods for gaze generalization.
The works in~\cite{bao2022generalizing,wang2022contrastive,cheng2022puregaze} are developed with a focus on domain adaptation for gaze estimation and encompass two-stage training schemes.
At the first stage, the three methods train feature invariant gaze networks that generalize well to variations of the respective feature.
In particular, RUDA~\cite{bao2022generalizing} trains a gaze estimation model with consistency between rotations of the same image, CRGA~\cite{wang2022contrastive} uses a contrastive loss to separate image features according to gaze and PureGaze~\cite{cheng2022puregaze} purifies face features to achieve higher gaze estimation performance.
The second stage of the above methods is focused on adapting the initially trained models to specific target domains.
As our method aims to train general gaze estimation models without knowledge of specific target domains, we implement the first-stage models of the above methods, namely RAT~\cite{bao2022generalizing}, CDG~\cite{wang2022contrastive} and PureGaze~\cite{cheng2022puregaze} and compare them with our proposed method in cross-dataset experiments.
To follow the evaluation protocol in the above works, we train all methods on EXG and G360 (alone or in conjunction with ITWG-MV) and test on MPII and GC.
\cref{tab:domain_gen} shows that the proposed method outperforms the baselines for gaze generalization when ITWG-MV is employed.
The compared methods are unable to regularize the pseudo-labels of ITWG-MV, while our method exploits them through $\mathcal{L}_{MV}$.


\begin{table}
\small
\setlength{\tabcolsep}{2.9pt}
\centering
\caption{\small Comparison with state-of-the-art in domain generalization for gaze estimation. In all experiments, our model outperforms the compared methods. Gaze error is in degrees (lower is better). Here IMV corresponds to ITWG-MV.}
\label{tab:domain_gen}
    \begin{tabular}{lcccccccc}
        \toprule
        & \multicolumn{2}{c}{\footnotesize EXG} & \multicolumn{2}{c}{\footnotesize EXG+IMV}
        & \multicolumn{2}{c}{\footnotesize G360} & \multicolumn{2}{c}{\footnotesize G360+IMV}
        \\
        \midrule
        Method & MPII & GC & MPII & GC & MPII & GC & MPII & GC 
        \\
        \midrule
        RAT~\cite{bao2022generalizing} &
        7.1 & \textbf{8.4} & 7.0 & 8.2 &
        9.3 & 9.0 & 9.1 & 8.5
        \\
        CDG~\cite{wang2022contrastive} &
        \textbf{6.7} & 9.2 & 6.9 & 9.5 &
        \textbf{7.0} & \textbf{8.3} & 8.1 & 8.9
        \\
        PureGaze~\cite{cheng2022puregaze} &
        7.9 & 8.7 & 7.7 & 9.3 &
        7.6 & 8.3 & 7.4 & 8.6
        \\
        \midrule
        3DGazeNet & 
        7.7 & 10.7 & \textbf{6.0} & \textbf{7.8} & 
        9.1 & 12.1 & \textbf{6.3} & \textbf{8.0}
        \\
        \bottomrule
    \end{tabular}
\end{table}

\begin{table}
\small
\setlength{\tabcolsep}{1.5pt}
\renewcommand{\arraystretch}{1}
\centering
\caption{\small Comparison between training targets Vector(V), Mesh(M) and Mesh+Vector(M+V) in within-dataset experiments (using only $\mathcal{L}_{GT}$). Training with target M+V leads to lower errors than state-of-the-art. Gaze error is in degrees (lower is better).}
\label{tab:gazeas3Drec}
   \begin{tabular}{ccccccccccc}
        \toprule
        Dataset & \multicolumn{7}{c}{Compared Methods} & \multicolumn{3}{c}{3DGazeNet}
        \\
        \midrule
          &\cite{o2022self}& \cite{cheng2022gazetr} &\cite{zhang2017s_mpiifgaze,cheng2021appearance}&\cite{Park2019ICCV_fewshot} & \cite{gaze360_2019,cheng2021appearance} & \cite{Kothari_2021_CVPR} & \cite{Zhang2020ETHXGaze} & V & M & M+V 
        \\
        \midrule
        MPII &4.04 &4.00&4.9&5.3  &4.06 & - & 4.8 & 4.1 & 4.2 & \textbf{4.0} 
        \\
        G360 &10.7&10.6&14.9& - & 11.1 & 10.1& - & 9.8 & 9.8 & \textbf{9.6} 
        \\
        GC   &-&- &-& 3.49 & - & - & 3.3 & 3.2 & 3.3 & \textbf{3.1} 
        \\
        EXG  &-&- &7.3& - & - & - & 4.5 & \textbf{4.2} & 4.4 & \textbf{4.2}
        \\
        \bottomrule
    \end{tabular}
\end{table}

\begin{table}
\small
\setlength{\tabcolsep}{1.8pt}
\renewcommand{\arraystretch}{0.8}
\centering
\caption{\small Comparison between training targets Vector and Mesh+Vector for domain generalization when employing our full model (\cref{eq:full_loss}). For the target Vector, we remove all mesh terms from the employed losses. In all experiments, the target Mesh+Vector results in a lower error. Gaze error is in degrees (lower is better).}
\label{tab:dense_vs_vector}
    \begin{tabular}{lcccccc}
        \toprule
        Training Dataset & Vector & Mesh+Vector & G360 & GC & EXG & MPII
        \\
        \midrule
        \multirow{2}{*}{ITWG-MV} & 
        \checkmark & - &         
        19.1 & 10.1 & \textbf{16.7} & 8.5               
        \\
        & - & \checkmark & 
        \textbf{18.1} & \textbf{9.0} & \textbf{16.7} & \textbf{7.6}  
        \\
        \midrule
        \multirow{2}{*}{G360+ITWG-MV} & 
        \checkmark & -          & 
        10.1 & 10.2 & 15.1 & 7.0  
        \\
        & - & \checkmark & 
        \textbf{9.3} & \textbf{8.0} & \textbf{14.6} & \textbf{6.3}  
        \\
        \midrule
        \multirow{2}{*}{GC+ITWG-MV} & 
        \checkmark  & - & 
        18.2 & 3.1 & 16.0 & \textbf{6.1}               
        \\
        & -          & \checkmark & 
        \textbf{17.6} & \textbf{3.0} & \textbf{15.5} & \textbf{6.1}  
        \\
        \midrule
        \multirow{2}{*}{EXG+ITWG-MV} & 
        \checkmark & - &
        16.5 & 10.2 & 4.5 & 6.6  
        \\
        & - & \checkmark &
        \textbf{15.4} & \textbf{7.8} & \textbf{4.3} & \textbf{6.0}
        \\
        \midrule
        \multirow{2}{*}{MPII+ITWG-MV} & 
        \checkmark & - & 
        17.8 & 8.2 & 15.2 & 4.8  
        \\
        & -          & \checkmark & 
        \textbf{17.6} & \textbf{6.8} & \textbf{14.9} & \textbf{4.2} 
        \\
        \bottomrule
    \end{tabular}
\end{table}

\subsection{Ablation studies}
\label{sec:ablation}

\noindent{\bf Gaze Estimation via 3D Eye Mesh Regression}
Here we experimentally evaluate our suggestion that gaze estimation benefits from replacing the training target from gaze vectors or angles to dense 3D eye coordinates.
To this end, we employ the fully supervised version of our model, utilizing data with exact ground truth and $\mathcal{L}_{GT}$ for training.
We conduct within-dataset experiments on MPII, GC, G360 and EXG for which specific  training-testing subsets are provided.
We compare against state-of-the-art methods~\cite{o2022self,cheng2022gazetr,zhang2017s_mpiifgaze,cheng2021appearance,Park2019ICCV_fewshot,gaze360_2019,Kothari_2021_CVPR,Zhang2020ETHXGaze} and report the results in \cref{tab:gazeas3Drec}.
In all cases our model performs better than the baselines, while combining the two modalities, i.e. dense 3D meshes and gaze vectors (M+V), improves performance compared to training with vector targets (V) or 3D mesh targets (M) alone.
This is possibly due to the distinct nature of the two modalities, i.e. the vectors provide exact label supervision, while meshes provide a robust representation which limits sparse prediction errors. 

We further evaluate the above suggestion in within- and cross-dataset experiments in G360, GC, EXG, and MPII after using the full version of our model trained with support from ITWG-MV.
For this experiment, we consider a version of our model in which all mesh terms are removed from the losses in \cref{eq:full_loss} (coined Vector) in an attempt to evaluate the contribution of meshes in our multi-view consistency framework.
From the results reported in \cref{tab:dense_vs_vector} we notice that in the majority of experiments employing combined training targets benefits performance.


\smallskip
\noindent{\bf The Effect of Gaze Pseudo-Labels and Multi-View Supervision}
Here we examine the contribution of our automatic geometry-aware pseudo-labels and the multi-view supervision loss of our approach. To this end, we consider three training scenarios which are the following: a) training with ITWG and its pseudo-labels as ground truth ($\mathcal{L}_{PGT}$), b) training with ITWG-MV utilizing only the multi-view consistency constraints and no pseudo-labels ($\mathcal{L}_{MV}$) and c) training with ITWG-MV while employing both pseudo-labels and the multi-view consistency loss ($\mathcal{L}_{PGT}$+$\mathcal{L}_{MV}$).
To further evaluate the effect of the pseudo-labels and multi-view loss, we repeat the above experiments by adding ground truth supervision from GC (+$\mathcal{L}_{GT}$).
We test our models on the test set of G360, GC, EXG, and MPII, and report the results in \cref{tab:ablation_mvloss}.
In all cases, combining our pseudo-labels and multi-view loss yields the lowest error in degrees.
It is worth noting that employing only pseudo-labels leads to better performance than only multi-view supervision.
This demonstrates that our 3D geometry-based method produces meaningful pseudo-labels, capable to support gaze-estimation generalization.
Lastly, training only with the multi-view loss on ITWG-MV leads to very high errors which is reasonable as no supervision for the eyeball topology exists, thus, the model outputs cannot follow the spherical shape of the eyeball template. 

\begin{table}
\small
\setlength{\tabcolsep}{2.3pt}
\centering
\caption{\small The effect of incorporating pseudo-ground truth and multi-view supervision during training. Both components contribute towards improving results in cross-dataset gaze estimation experiments. Gaze error is in degrees (lower is better).}
\label{tab:ablation_mvloss}
    \begin{tabular}{lccccccc}
        \toprule
        Dataset & $\mathcal{L}_{GT}$ & $\mathcal{L}_{PGT}$ & $\mathcal{L}_{MV}$ & G360 & GC & EXG & MPII
        \\
        \midrule
        ITWG       & -          & \checkmark & -          & 23.1 & 14.8 & 24.3 & 13.6
        \\
        ITWG-MV    & -          & -          & \checkmark & 47.4 & 33.2 & 41.1 & 32.8
        \\
        ITWG-MV    & -          & \checkmark & \checkmark & \textbf{18.1} & \textbf{9.0} & \textbf{16.7} & \textbf{7.6}  
        \\
        \midrule
        GC         & \checkmark& -          & -          & 27.5 & 3.1 & 28.4 & 10.4
        \\
        GC+ITWG    & \checkmark& \checkmark & -          & 21.4 & 3.2   & 23.7 & 9.1
        \\
        GC+ITWG-MV & \checkmark& -          & \checkmark & 24.7 & 3.5   & 26.2 & 10.1
        \\
        GC+ITWG-MV & \checkmark& \checkmark & \checkmark & \textbf{17.6} & \textbf{3.0} & \textbf{15.5} & \textbf{6.1}  
        \\
        \bottomrule
    \end{tabular}
\end{table}


\smallskip
\noindent{\bf The Effect of Head Pose Distribution of ITWG}
Head pose distribution difference between the train and test set is one of the main reasons that gaze-estimation models fail in cross-dataset situations.
To close the gap between different training and testing scenarios, we have designed ITWG, a large-scale dataset with widespread variation in head pose and gaze angles.
To study the effect of the head pose variation of ITWG in our experiments, we employ different subsets of ITWG with various levels of head pose variation and conduct cross-dataset experiments with them.
In particular, we consider four subsets of ITWG, with maximum yaw angles of $5^o$, $20^o$, $40^o$ and $90^o$ (all) respectively.

\begin{figure}
\centering
    \includegraphics[width=1\linewidth]{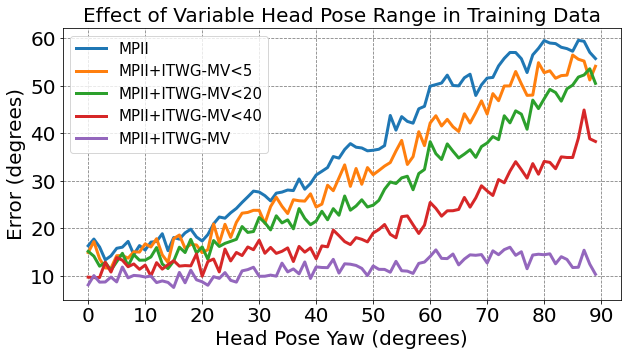} 
\caption{\small Gaze error of G360 across head poses when training with MPII and subsets of ITWG-MV. Wider range of head poses in the ITWG-MV data, lead to significantly lower errors in large poses.}
\label{fig:ablation_headpose}
\end{figure}

We train 3DGazeNet with ground truth supervision from MPII as well as pseudo-labels and multi-view supervision from the four versions of ITWG-MV.
The results of testing on G360 are presented in \cref{fig:ablation_headpose}. 
The resulting curves clearly demonstrate the effect of the available head pose variation in the training data. 
Specifically, utilizing the entirety of ITWG-MV leads to the lowest errors which are relatively consistent across the head pose range.
As expected, decreasing the available head pose variation, increasingly affects model performance with the worst case being training with MPII alone.
Based on the above finding we argue that the gap between small and wide distribution gaze datasets (regarding head pose) can effectively close by employing similarly large distribution unlabeled face datasets, which is crucial for training plug-n-play gaze estimation models that can be directly employed in applications.

%% file: sec/5_conclusion.tex
\section{Limitations and Conclusion}
\label{sec:limitations}

In \cref{sec:experiments}, we shown that pseudo-ground truth can be effectively utilized in gaze estimation.
In our experimentation, this usually resulted in weaker labels in the pitch axis, which is possibly due to the limited vertical variation of gaze in the data, or flaws of our pseudo-annotation pipeline. 
Moreover, pseudo-annotation accuracy is related to the accuracy of 3D face and 2D iris alignment.
Lastly, our current method cannot operate on images without a visible face (when the face is looking away from the camera). 


In this work, we present a novel weakly-supervised method for gaze generalization, based on dense 3D eye mesh regression.
We demonstrate that by utilizing both 3D eye coordinates and gaze labels during training, instead of just gaze labels, we can achieve lower prediction errors.
Moreover, we explore the possibility of exploiting the abundantly available in-the-wild face data for improving gaze estimation generalization.
To that end, we propose a novel methodology to generate robust, 3D geometry-aware pseudo ground truth labels, as well as a multi-view weak-supervision framework for effective training.
By enforcing these constraints, we are able to successfully utilize in-the-wild face data and achieve improvements in cross-dataset and within-dataset experiments.

%% file: sec/X_suppl.tex
\clearpage
\setcounter{page}{1}
\maketitlesupplementary

\section{Model Demo}

To showcase the effectiveness of our method in real conditions, we have trained a general gaze estimation model based on 3D eye mesh regression.
The model has been trained with the four gaze datasets discussed in this paper (MPIIFaceGaze~\cite{zhang15_cvpr_mpiigaze,zhang2017s_mpiifgaze}, Gaze360~\cite{gaze360_2019}, ETH-XGaze~\cite{Zhang2020ETHXGaze} and GazeCapture~\cite{cvpr2016_gazecapture}), as well as the in-the-wild images with pseudo-labels from our ITWG-MV dataset which we employ within our proposed framework.
Combining the accuracy of gaze datasets and the diversity of ITWG-MV, our model is robust to a wide variety of scenarios, e.g. challenging head poses, lighting settings and occlusions, that are commonly encountered in real scenarios.
A live demo of our method can be found \href{http://35.92.43.43:7011/}{at this link}. 

\section{Implementation Details}

\noindent\textbf{Data Augmentation} As model input we center crop 3 patches (left eye, right eye and face) resize them to shape $128 \times 128 \times 3$ and stack them channel-wise.
Before stacking the image patches we randomly scale, translate, flip, and add noise on the color channel with probability 0.5.
All geometric augmentations are also applied to the coordinates of eyes given as ground truth or pseudo-ground truth.
Due to different image quality on each dataset we also add Gaussian blur to the images.
The intensity of the blur is randomly selected from a set of kernels.
Note that we augment each patch with the same augmentation parameters. 

\medskip
\noindent\textbf{Training Details} We train our method using a Adam optimizer (weight decay at 0.0005, and batch size of 128) on a single Tesla V100-PCIE (32GB) GPU.
The learning rate starts from 1e-6, linearly warming up to 1e-4 in the first 3 epochs and then divided by 10 at 60 and 80 epochs.
The training process is terminated at 100 epochs. 

\medskip
\noindent\textbf{2D Iris Landmark Localization} To localize 2D iris landmarks from images we started by using the model from~\cite{Park2018ETRA}.
Even though the available pre-trained models of~\cite{Park2018ETRA} perform well on high resolution datasets, such as ETH-XGaze, they perform poorly on images with low resolution, such as the ones from Gaze360, AVA and CMU datsets. 
To overcome this problem we trained a version of our mesh-based model using predictions of~\cite{Park2018ETRA} on high resolution images.
In fact, we first applied~\cite{Park2018ETRA} on ETH-XGaze and FFHQ datasets which resulted on 2D iris landmarks for these images.
Then, using our pseudo-label generation pipeline (described in Sec. 3.2 of the paper, fig. 3(c)) we fit 3D eyeballs on these images. 
Next, we trained a 3D eyeball mesh reconstruction model, with the same architecture as the one described in the paper but without including the gaze head. 
Because of the data augmentation we applied during training (as described above), this model performs well on images with low resolution, occlusions and low-light conditions and gives us reliable 2D iris landmarks to either fit 3D eye meshes on existing gaze datasets using ground truth (fig. 3(b) of the paper) and generate psuedo-labels (fig. 3(c) of the paper).
Note that we only use this model to acquire 2D iris landmarks and do not care about its performance in gaze estimation.

\begin{figure}[t]
\centering
\includegraphics[width=1.\linewidth]{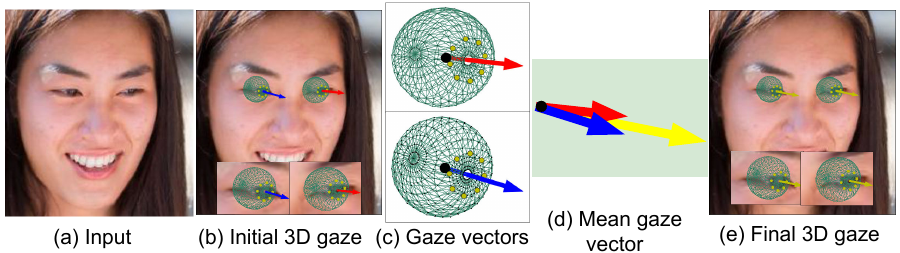}
\caption{Calculating 3D gaze from eye meshes. Given 3D eye meshes extracted by our method, we calculate gaze direction as the mean of the two independent gaze vectors from the left and right eyes.}
\label{fig:3Dgaze}
\end{figure}

\medskip
\noindent\textbf{Calculating Gaze Direction from 3D Eye Meshes}
In this work we have proposed a method to estimate 3D eye meshes from images and employ them for gaze estimation.
Having recovered a 3D eyeball mesh with topology adhering to our 3D eyeball template, we calculate gaze from the orientation of the central axis of the eyeball.
Particularly, we calculate 3D gaze vectors using the centre of the eyeball and the centre of the iris as shown in \cref{fig:3Dgaze}(c).
After obtaining 3D gaze vectors from both left and right eyes, we add and normalize the two vectors to retrieve a mean eyeball-based gaze prediction, \cref{fig:3Dgaze}(d).
Lastly, we add and normalize the mean eyeball-based gaze prediction with the gaze vector predicted by the gaze head of our model.
This vector constitutes the final gaze output of our model.

\begin{figure}[t]
\centering
    \includegraphics[width=0.25\textwidth]{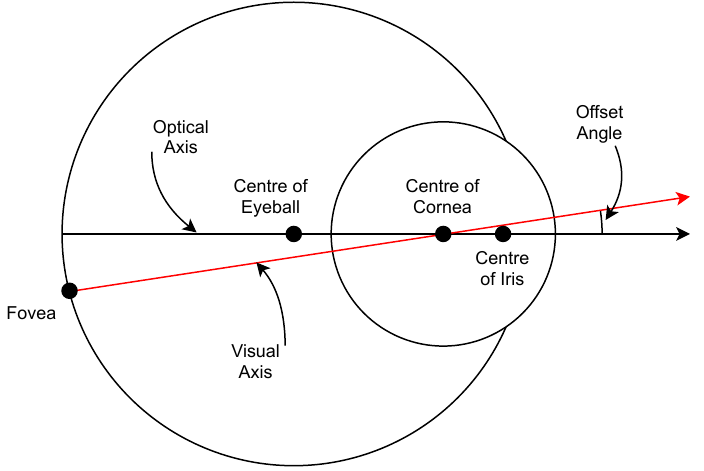}
    \caption{Eyeball anatomy demonstrating the offset between the optical and visual axis.}
\label{fig:eye_anatomy}
\end{figure}

Using the central axis of the 3D eye meshes to calculate the gaze direction is a reasonable approximation for our model.
In practice, as seen in \cref{fig:eye_anatomy}, an offset angle (the kappa coefficient) exists between the central (optical) and visual axes of eyes, which is subject dependent and varies between $-2^o$ to $2^o$ across the population~\cite{Yu_2019_CVPR}.
Even though accounting for this offset is crucial for person specific gaze estimation~\cite{He_2019_ICCVW,Liu2018ADA,Park2019ICCV_fewshot,Yu_2019_CVPR}, in the case of cross-dataset and in-the-wild gaze generalization, errors are much larger than the possible offset as can be seen from our all our cross-dataset experiments.
Therefore, in that case data diversity is more important than anatomical precision.
Besides, by combining the results of the eye mesh and gaze head of our model, we benefit both from the robustness offered from dense coordinate prediction and the accuracy of directly predicting gaze labels.
This is supported by the results of Sec. 4.3 of the paper.

\section{Application on Gaze Redirection}

One of the main objectives of this work is to train gaze tracking models that generalize well to unseen domains and in-the-wild conditions.
In this way models can be employed in a plug-n-play fashion by applications, offering reliable gaze predictions without the need for knowledge of a specific target domain or parameter fine-tuning.
As a practical use case, we test the effect of 3DGazeNet in the tasks of gaze redirection and gaze correction (gaze redirection to the camera, i.e. $(0^o, 0^o)$ pitch and yaw angles).
To that end, we employ VFHQ~\cite{xie2022vfhq} a high resolution video dataset of talking faces, initially built for video-based face super-resolution, which contains 16K clips and a wide variation of environments, illumination conditions and subject nationalities.

We design the gaze redirection experiment as follows.
First, we split VFHQ in a training and a test set containing 12K and 4K videos each, and sample a total of 200K and 20K frames from each subset.
Then, we extract gaze labels using 3DGazeNet trained on all datasets (MPII, GC, EXG, G360, ITWG-MV) and employ them for training an image translation network as in~\cite{he2019photo}.
For clarity in comparisons we name the image translation network GazeRet-ITW.
To evaluate the above model we generate 5 images with random gaze labels in the range $[-50^o, 50^o]$ yaw and $[-30^o, 30^o]$ pitch from each image of the test set. We additionally generate an image with gaze direction $(0^o, 0^o)$.
We predict the gaze of the translated images using the initial gaze estimation model and measure the redirection error between the predictions and the known target labels.

\begin{figure*}
\centering
    \includegraphics[width=\linewidth]{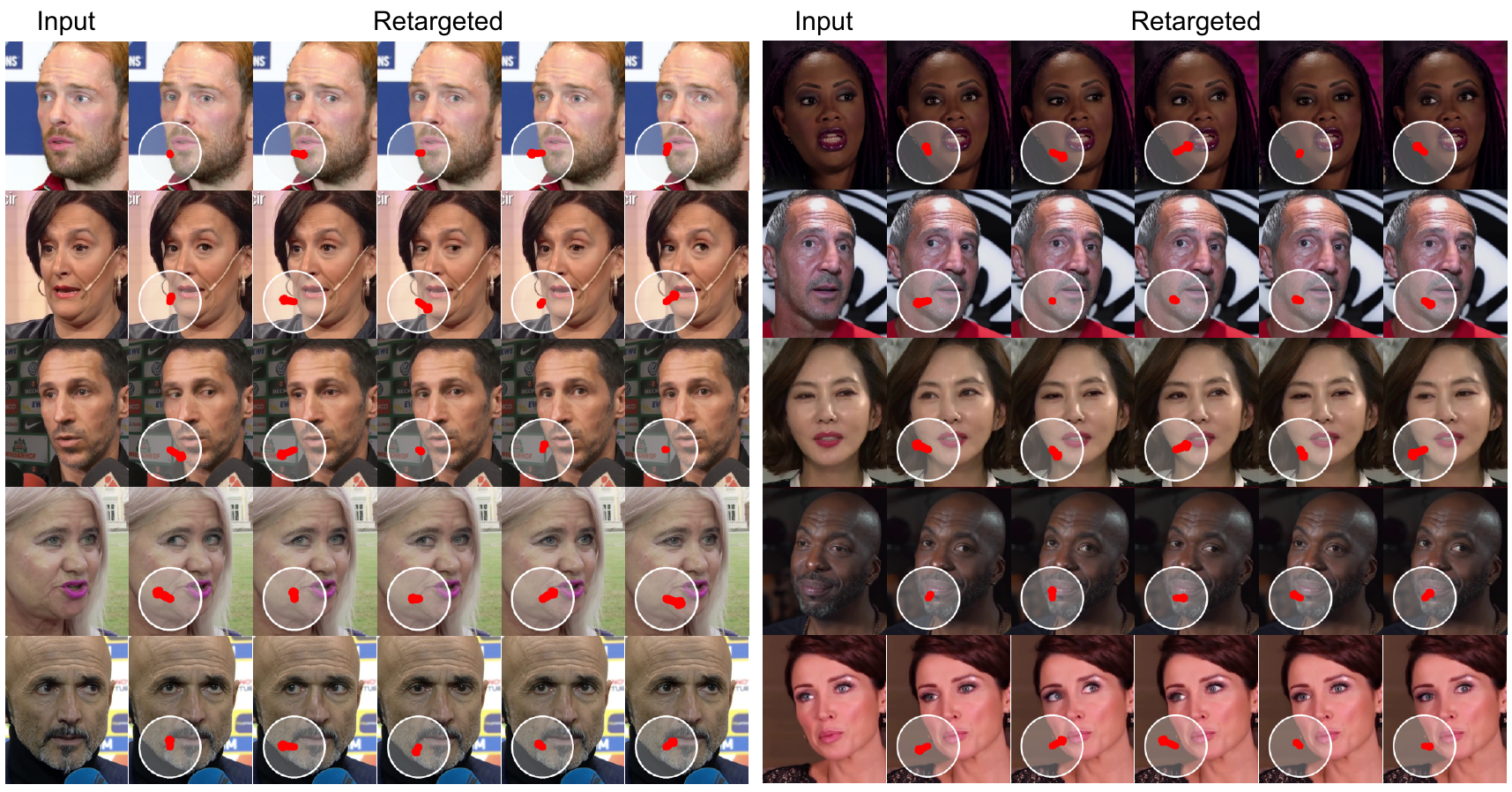}
    \caption{\small Gaze redirection on images of VFHQ. The redirected images have been produced by GazeRet-ITW. The red arrow at the bottom left of each redirected image indicates the target gaze label used to generate that particular image.}
\label{fig:retimages}
\end{figure*}

To highlight the benefits of 3DGazeNet in the above task, we repeat the experiment using VFHQ gaze labels extracted from a model trained with all gaze datasets except for ITWG-MV (thus, using only ground truth supervision) which we name 3DGazeNet-GT. Also, we name this version of the image translation network as GazeRet-GT.
\cref{tab:reterror} presents the redirection errors in the above two cases, as well as in cross-evaluation scenarios, for both redirection and correction.
In all cases, images produced with GazeRet-ITW lead to lower errors showcasing the benefits of augmenting gaze datasets with ITWG-MV and weakly-supervising training with our multi-view consistency loss. 
This is due to the fact that 3DGazeNet can produce reliable gaze labels in the unseen domains of the VFHQ video clips, while the simple gaze tracking model cannot.
Images with manipulated gaze direction using GazeRet-ITW can be seen in \cref{fig:retimages}.

\noindent{\bf Implementation details} For image-to-image translation we adapt the model architecture from StarGANv2~\cite{choi2020stargan}, removing the style encoder and the latent-to-style mapping network, and replacing the style codes with gaze vectors as conditions to the generator. 
We also adjust the discriminator to predict continuous gaze labels as in ~\cite{pumarola2018ganimation,ververas2020slidergan}, inducing precise gaze in image translation.
Similarly to the above models, we operate using a cyclic reconstruction loss and a gaze prediction loss, without image pairs.
As model input, we crop image patches of size 256$\times$512 containing both eyes and perform image translation on them.
We sample target gaze labels (generator conditioning labels) from the training set, as well as randomly.
During testing, we blend the full face input images with the output patches.
Lastly, we train the model with learning rate 0.0001 and batch size 64, using Adam optimizer for a total of 300K steps.

\begin{table}
\centering
\caption{\small Gaze Redirection error on images of our test split of VFHQ. Images generated with the GazeRet-ITW method lead to lower errors when evaluated by either of the gaze estimation models for both redirection and correction. The gaze errors are in degrees (lower is better).}
\label{tab:reterror}
\begin{tabular}{lcccc}
    \toprule
    Method & 
    3DGazeNet-GT & 3DGazeNet
    \\
    \midrule
     &
    \multicolumn{2}{c}{\small Redirection}
    \\
    \midrule
    GazeRet-GT & 
    9.4 & 10.1 
    \\
    GazeRet-ITW & 
    \textbf{6.7} & \textbf{8.4}
    \\
    \midrule
     & 
    \multicolumn{2}{c}{\small Correction}
    \\
    \midrule
    GazeRet-GT & 
    7.1 & 7.6
    \\
    GazeRet-ITW & 
    \textbf{3.8} & \textbf{5.2}    
    \\
    \bottomrule
\end{tabular}
\end{table}

\section{Comparison with Model-Based Methods}

As 3DGazeNet infers gaze based on 3D models of the eyes, it is sensible to mention related methods that use shape representations of the eyes. In particular, CrtCLGM+MTL~\cite{Yu_2018_ECCVW} learns to predict gaze simultaneously with sparse 2D landmarks of the eye region and find their multi-task objective to be beneficial in comparison to just predicting gaze.
CrtCLGM+MTL achieves $5.7^o$ error on the within-dataset experiment on UTMV dataset~\cite{sugano2014utmv} (which includes 64K images and 50 subjects), while our method trained just with ground truth meshes and gaze vectors achieves $5.5^o$, following the same experimental settings as~\cite{Yu_2018_ECCVW}.
Another related method, DPG~\cite{Park_2018_ECCV_dpge}, learns 2D segmentation maps of the eyeball and iris and employ those to infer gaze.
DPG achieves error $4.5^o$ on MPII and $3.6^o$ on Columbia~\cite{CAVE_Columbia2013} (5880 images and 56 subjects) on within-dataset experiments, while our method achieves $4.0^o$ and $3.1^o$ respectively. 

Other than the above methods which regress gaze as an output of a network,~\cite{Park2018ETRA} and~\cite{3dmm_wood2016} infer gaze using reconstructions of the 3D eyeball, similarly to our method. In particular,~\cite{Park2018ETRA} infers 2D landmarks of the eye region as well as the eyeball center and radius, which allows the reconstruction of a 3D eyeball and thus the prediction of a 3D gaze vector. \cite{Park2018ETRA} is trained on rendered images using a 3D model of the eye region and achieves error $7.1^o$ on the Columbia dataset without any annotation from it. Our method trained only with ITWG achieves error $5.6^o$ on the same dataset. Moreover,~\cite{3dmm_wood2016} have proposed a parametric model fitting approach in which they fit a 3D morphable model of the eye region and eyeball shape, texture and illumination and infer gaze based on the 3D eyeballs. Their model-fitting approach has given $7.5^o$ error on the Columbia dataset.

\begin{table}[t]
\small
\centering
\caption{\small The effect of head pose variation of ITWG. Starting from either MPII or GC ground truth datasets, which include much smaller head pose and gaze variation than G360, incorporating data from ITWG with increasingly more diverse head pose, leads to lower gaze error (measured in degrees, lower is better). The error decreases significantly even in the case of G360 $<5^o$ which indicates that the pseudo-labels of ITWG are meaningful and its face and environmental variation useful.}
\label{tab:ablation_itwg}
    \begin{tabular}{lcccc}
        \toprule
        Training Datasets & \multicolumn{4}{c}{Gaze360 Test Subsets}
        \\
        \midrule
        & $<5^o$ & $<20^o$ & $<40^o$ & $<90^o$
        \\
        \midrule
        MPII                & 35.6 & 27.2 & 24.9 & 25.7
        \\
        MPII+ITWG-MV$<5^o$  & 33.0 & 26.6 & 23.7 & 22.9
        \\
        MPII+ITWG-MV$<20^o$ & 27.2 & 23.1 & 21.2 & 20.3
        \\
        MPII+ITWG-MV$<40^o$ & 24.7 & 21.4 & 20.0 & 19.5
        \\
        MPII+ITWG-MV all    & \textbf{22.3} & \textbf{20.2} & \textbf{18.9}  & \textbf{17.6}
        \\
        \midrule
        GC                & 36.4 & 28.5 & 25.2 & 27.5
        \\
        GC+ITWG-MV$<5^o$  & 29.9 & 24.2 & 22.1 & 23.1
        \\
        GC+ITWG-MV$<20^o$ & 27.4 & 22.8 & 20.8 & 20.7
        \\
        GC+ITWG-MV$<40^o$ & 25.8 & 21.8 & 19.9 & 19.8
        \\
        GC+ITWG-MV all    & \textbf{23.3} & \textbf{20.4} & \textbf{19.1} & \textbf{17.6}
        \\
        \bottomrule
    \end{tabular}
\end{table}

\begin{figure}[t]
\setlength{\tabcolsep}{0pt}
\centering
    \begin{tabular}{cc}
         \includegraphics[width=.53\linewidth]{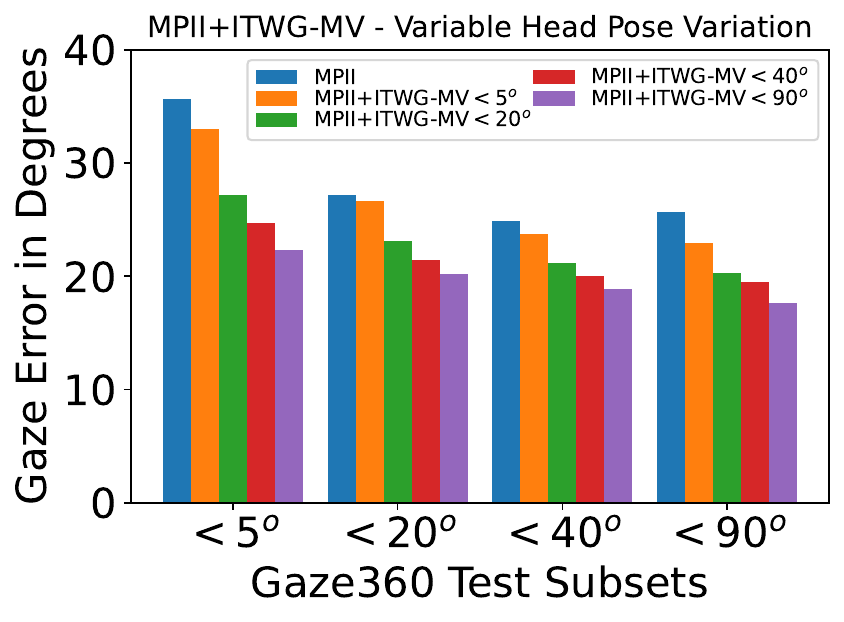}
         &
         \includegraphics[width=.46\linewidth]{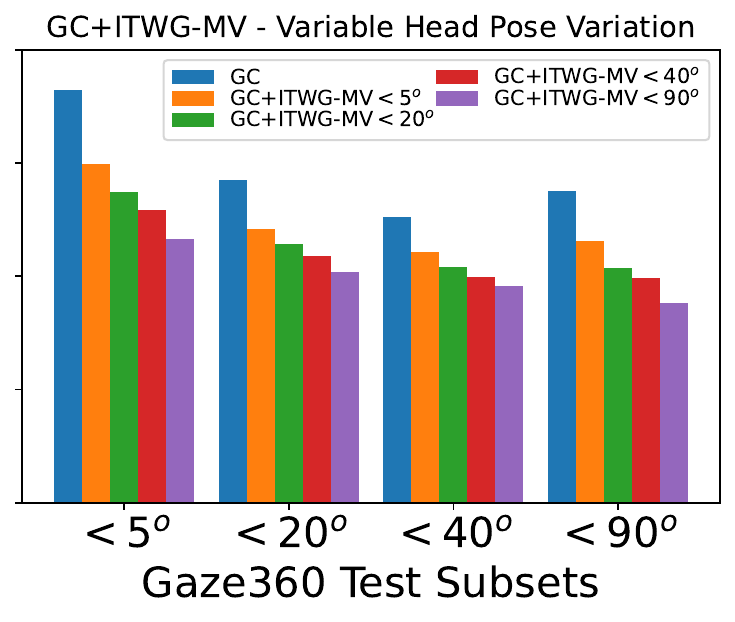}
         \\
         (a) MPII & (b) GC
    \end{tabular}
\caption{\small Bar plots showing the effect of head pose variation of ITWG in cross-dataset experiments for the case of training with ground truth from (a) MPII and (b) GC and testing on G360. The data are taken from \cref{tab:ablation_itwg}.}
\label{fig:itwg_headpose}
\end{figure}

\section{Additional Ablation on the effect of ITWG's Head Pose Variation}

In this section we present further analysis on the effect of head pose variation of ITWG on the model's generalization ability.
For training we employ different subsets of ITWG based on head pose ($<5^o$, $<20^o$, $<40^o$, $<90^o$(all)) and additional ground truth supervision from MPIIFaceGaze (MPII) or GazeCapture (GC).
We present results of our models for different subsets of G360's test set, based again on the head pose yaw angle ($<5^o$, $<20^o$, $<40^o$, $<90^o$).
Results reported in \cref{tab:ablation_itwg} and \cref{fig:itwg_headpose} demonstrate that an improvement of $6^o$ to $13^o$ (24\% to 37\%) is achieved in all cases (for all subsets of G360) between the baselines of training just with gaze datasets (MPII, GC) and our full method of including ITWG and multi-view supervision in training.
It is also worth noticing that performance consistently increases when training with more diverse subsets of ITWG in all cases.
Lastly, it is worth noticing that even though performance increase is expected for the subsets of G360 with large head pose values ($>40^o$), as MPII and GC do not include such images at all, the larger increase in performance is seen for near-frontal images.
This fact, validates the effectiveness of our pseudo-labelling method and our multi-view supervision algorithm.

\section{Additional Ablation on the effect of Pseudo-Labels and Multi-View Supervision}

In this section we present further evaluations on the effect of our pseudo-labels and multi-view consistency constraints during training.
To that end, we repeat the experiments described in Sec. 4.3 of the main paper for 3 additional cases of ground truth supervision and report the results in \cref{tab:ablation_mvloss_supp}.
In particular, we utilize G360, EXG and MPII as source datasets with valid ground truth.
From the results we can draw the conclusion that our method is always effective when there are large differences between the source and target dataset head pose and gaze variation. 
In such cases ITWG helps to close the gap and results in significant improvement.
Only small improvement is noticed for the within-dataset experiment on Gaze360, while results do not improve for within-dataset experiments on EXG and MPII.
EXG has been captured in lab conditions and already includes a wide variety of head poses and gaze directions, thus augmenting our model's training with more in-the-wild data from ITWG does not benefit within dataset evaluation.
Similarly for MPII, which is very restricted in terms of environments and pose variation, the best performance is achieved with ground truth supervision only.

\begin{table}
\small
\setlength{\tabcolsep}{3pt}
\centering
\caption{\small The effect of incorporating pseudo-ground truth and multi-view supervision during training. Both components contribute towards improving results in cross-dataset gaze estimation experiments. Gaze error is in degrees (lower is better).}
\label{tab:ablation_mvloss_supp}
    \begin{tabular}{lcccccc}
        \toprule
        Dataset & $\mathcal{L}_{PGT}$ & $\mathcal{L}_{MV}$ & G360 & GC & EXG & MPII
        \\
        \midrule
        G360         & -          & -          & 9.6 & 12.1 & 18.3 & 9.1
        \\
        G360+ITWG    & \checkmark & -          & 9.4 & 10.2 & 16.4 & 8.2
        \\
        G360+ITWG-MV & -          & \checkmark & 9.6 & 11.7 & 18.1 & 9.1
        \\
        G360+ITWG-MV & \checkmark & \checkmark & \textbf{9.3} & \textbf{8.0} & \textbf{14.6} & \textbf{6.3}  
        \\
        \midrule
        EXG         & -          & -          & 22.1 & 10.7 & \textbf{4.3} & 7.7
        \\
        EXG+ITWG    & \checkmark & -          & 19.4 & 11.1 & 4.7 & 6.8
        \\
        EXG+ITWG-MV & -          & \checkmark & 21.6 & 12.0 & 4.5 & 7.4
        \\
        EXG+ITWG-MV & \checkmark & \checkmark & \textbf{15.4} & \textbf{7.8} & \textbf{4.3} & \textbf{6.0}  
        \\
        \midrule
        MPII         & -          & -          & 23.6 & \textbf{6.3} & 26.3 & \textbf{4.0}
        \\
        MPII+ITWG    & \checkmark & -          & 19.8 & 7.2 & 21.7 & 4.4
        \\
        MPII+ITWG-MV & -          & \checkmark & 22.9 & 7.4 & 25.1 & 4.2
        \\
        MPII+ITWG-MV & \checkmark & \checkmark & \textbf{17.6} & 6.8 & \textbf{14.9} & 4.2  
        \\
        \bottomrule
    \end{tabular}
\end{table}

\begin{figure*}[t]
\centering
    \includegraphics[width=1.\linewidth]{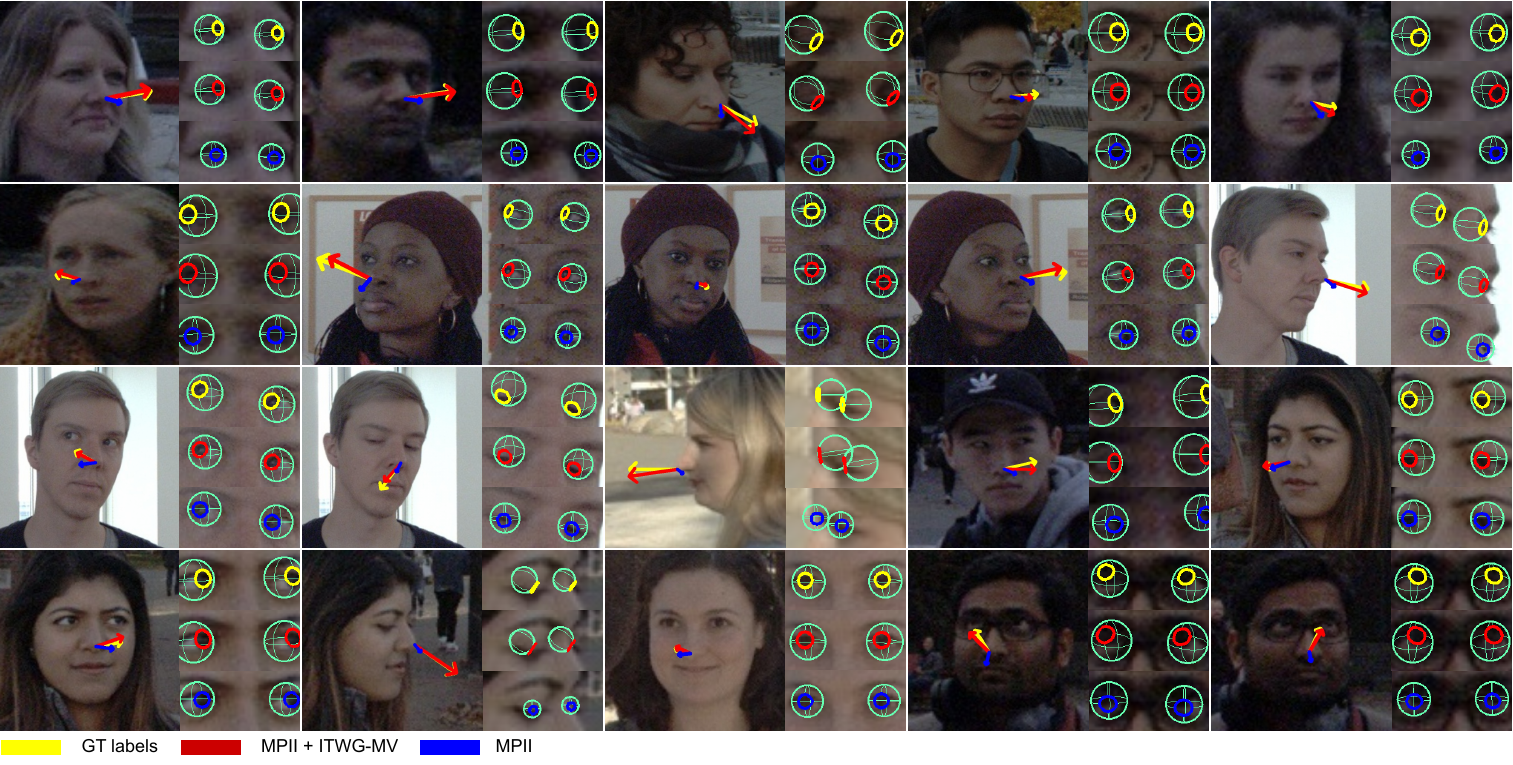}
    \caption{Results of our models trained with MPII (blue vectors) and combined MPII and ITWG with multi-view supervision (red vectors), applied on the test set of G360 (yellow vectors). The predicted gaze directions are closer to the ground truth when ITWG is included in training. Especially for side and profile views, the effect of the pseudo-labels is significant.}
\label{fig:mpii_to_g360}
\end{figure*}

\begin{figure*}[t]
\centering
    \includegraphics[width=1.\linewidth]{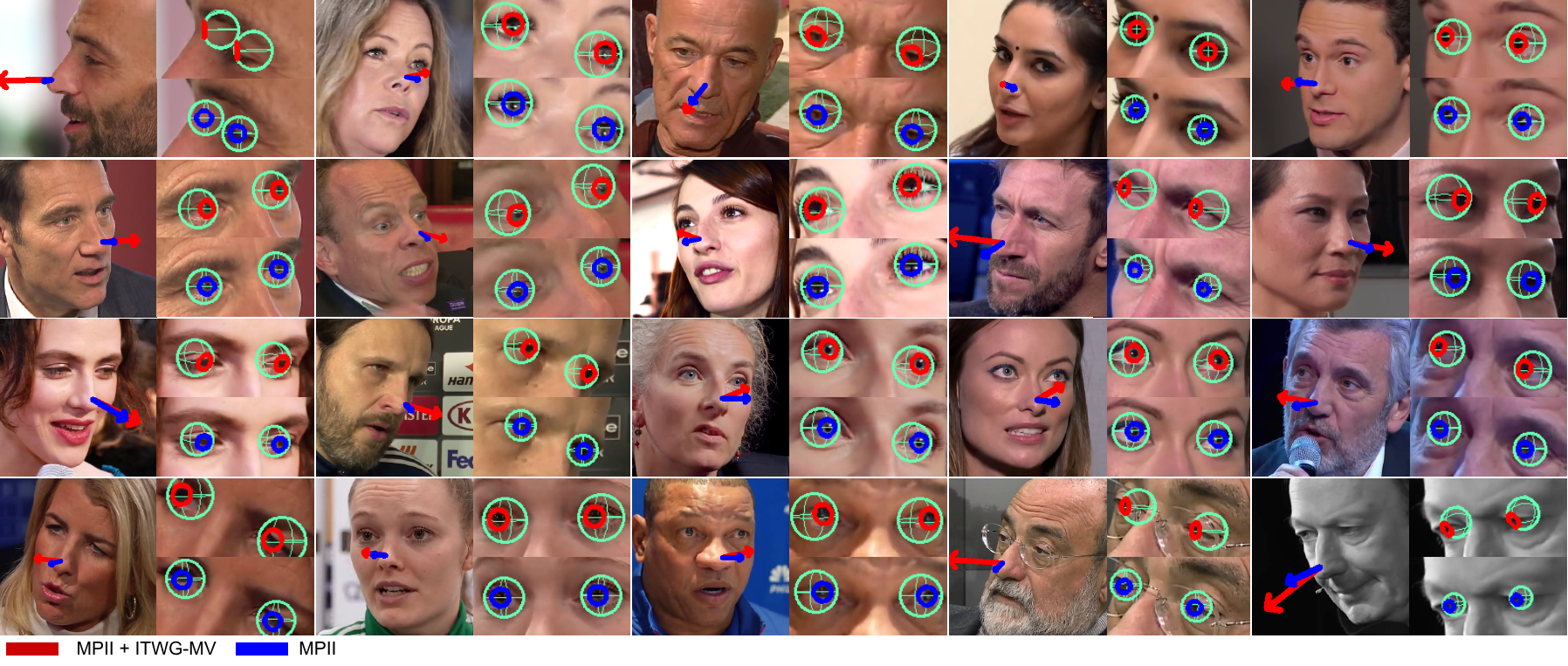}
    \caption{Results of our models trained with MPII (blue vectors) and combined MPII and ITWG with multi-view supervision (red vectors), applied on images of VFHQ. Our full model predicts robust gaze labels across all head pose angles, especially for profile ones the effect of the pseudo-labels is significant.}
\label{fig:mpii_to_itw}
\end{figure*}

\begin{figure*}[t]
\centering
\includegraphics[width=1.\linewidth]{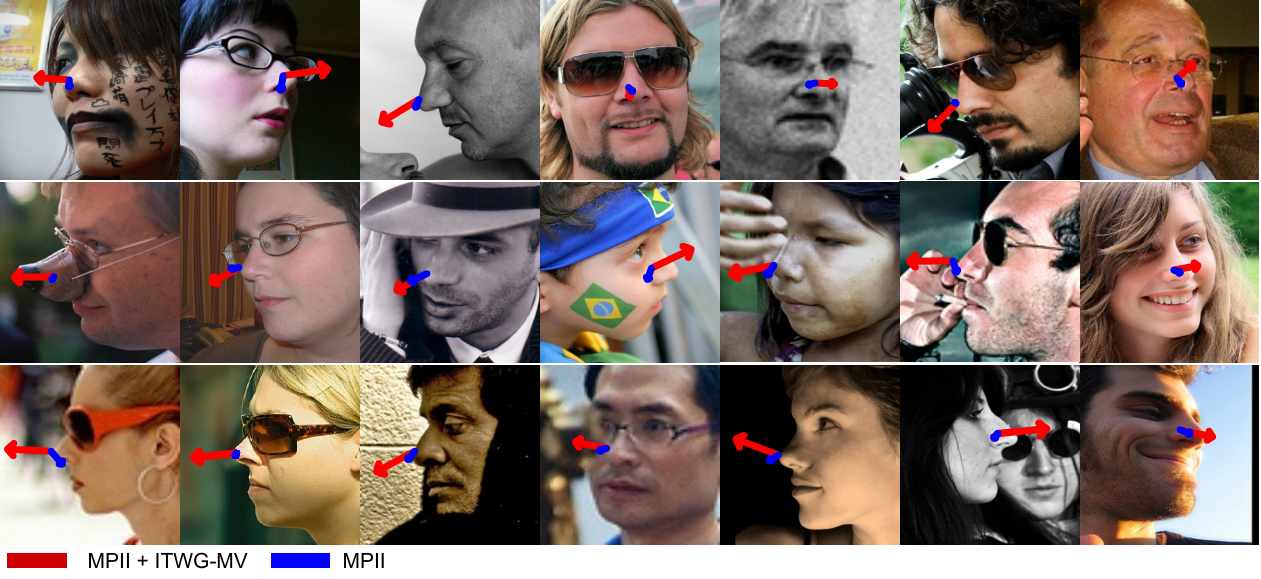}
\caption{Results from applying our model on difficult cases including faces in profile pose, faces with glasses and faces with occlusions or low resolution. Our model can successfully handle these difficult scenarios and produce reliable gaze predictions.}
\label{fig:difficult}
\end{figure*}

\section{Qualitative Results for 3D Gaze Estimation}

Here we visualize gaze predictions of our model for training scenarios discussed in Sec. 4.3 of the main paper (The Effect of Head Pose Distribution of ITWG).
In particular, we present the results of our model in the two edge cases of Fig. 6 of the paper, i.e. a) only MPII is employed for training and b) MPII and the whole ITWG with multi-view supervision.
Testing is performed on the images of G360. \cref{fig:mpii_to_g360} includes results for the two cases as well as the ground truth labels of G360.
As can be seen, the predicted gaze directions are much closer to the real ones when in-the-wild face data from ITWG are employed for supervision. Especially for profile views, the effect of the pseudo-labels is significant.

To further evaluate our method, we apply the above two models on arbitrary in-the-wild face images (images of VFHQ and AFLW) and present the results on \cref{fig:mpii_to_itw} and \cref{fig:difficult}.
As actual gaze accuracy cannot be measured for such images (ground truth data are not available), we attempt to draw conclusions based on observation.
From the visualizations it can be seen that for side and profile views, our multi-view supervision method (MPII + ITWG-MV) performs substantially better, while for near-frontal ones the predictions improve. 

Especially, in \cref{fig:difficult} we have included cases which we consider difficult for gaze estimation.
These include images in profile views, images with occluded eyes or eye glasses as well as images with low resolution and bad illumination.
Results demonstrate that 3DGazeNet can produce reliable gaze labels for all of the above cases, which makes it ideal for real applications operating in unrestricted environments.